\documentclass{article}

\usepackage{arxiv}

\usepackage[utf8]{inputenc} 
\usepackage[T1]{fontenc}    
\usepackage{hyperref}       
\usepackage{url}            
\usepackage{booktabs}       
\usepackage{amsfonts}       
\usepackage{nicefrac}       
\usepackage{microtype}      
\usepackage{lipsum}
\usepackage{graphicx}
\usepackage{amsmath}
\graphicspath{ {./images/} }
\usepackage{graphicx}
\usepackage{subcaption}
\usepackage{placeins}
\usepackage{chngcntr}

\title{CapCLIP: A Vision-Language Representation Alignment Approach for Wireless Capsule Endoscopy Analysis}

\author{
Haroon Wahab\thanks{Code will be made publicly available after acceptance.} \\
  School of Computer Science, AI and Electronics\\
  Faculty of Engineering and Digital Technologies
  University of Bradford\\
  Bradford, UK \\
  \texttt{m.h.wahab@bradford.ac.uk} \\
   \And
 Irfan Mehmood \\
School of Management \\
Faculty of Mgmt, Law \& Social Sciences\\
University of Bradford\\
Bradford, UK\\
  \texttt{i.mehmood4@bradford.ac.uk} \\
  \And
 Hassan Ugail \\
  Centre for Visual Computing and Intelligent Systems \\
  University of Bradford\\
  Bradford, UK \\
  \texttt{h.ugail@bradford.ac.uk} \\
}

\begin{document}
\maketitle
\begin{abstract}
 Wireless capsule endoscopy (WCE) enables non-invasive visual assessment of the small bowel, but its clinical utility is constrained by the large volume of frames generated per examination and the difficulty of recognising subtle abnormalities under highly variable imaging conditions. Existing learning-based approaches for WCE are predominantly vision-only, often confined to narrow pathology sets, and show limited transfer across datasets and centres. To address these limitations, this study introduces CapCLIP, a domain-specific vision-language representation learning framework for WCE. CapCLIP aligns capsule endoscopy frames with clinically grounded textual descriptions derived from standardised nomenclature and pathology-aware caption templates, thereby learning embeddings that are both semantically informed and transferable. The proposed framework is evaluated against relevant open-source vision and vision-language foundation models under strict zero-shot conditions using unseen WCE datasets. Evaluation covers three downstream tasks: K-nearest neighbour classification, CLIP-style image-text classification, and text-to-image retrieval. Across these settings, CapCLIP consistently outperforms the compared baselines, with particularly strong gains in zero-shot image-text classification and cross-modal retrieval on out-of-distribution datasets. The results indicate that language-guided representation learning can improve both generalisation and semantic interpretability in WCE analysis. These findings position CapCLIP as a step toward foundation models tailored to capsule endoscopy and support the use of language-grounded WCE analysis.
\end{abstract}


\section{Introduction}
Wireless capsule endoscopy (WCE) is an established modality for small-bowel examination, offering minimally invasive visualisation of regions that are difficult to access using conventional endoscopic procedures \cite{iddan2000swallowing}. Its utility is limited, however, by the large volume and complexity of the data it produces: a single study may contain tens of thousands of frames, many of them redundant, while clinically relevant abnormalities can be subtle and transient. Consequently, manual review remains time-consuming and reader-dependent, motivating computational methods that can reduce review burden without compromising clinically meaningful findings.

Despite the ongoing research on computer-aided WCE analysis, most existing models remain constrained by limited generalisability and narrow evaluation settings. Many are designed for single abnormalities or predefined label sets, and their reported gains are commonly tied to retrospective experiments with restricted dataset diversity. Such models may therefore perform well within a benchmark yet remain untested under cross-dataset heterogeneity such as pathology prevalence, acquisition conditions, and image appearance \cite{wahab2023}. In addition, their outputs are usually limited to fixed class predictions, with interpretability relying largely on post-hoc visualisation rather than semantically grounded reasoning.

Overcoming the limitations of dataset-bound WCE models requires a representation learning scheme that promotes transferable structure in the learned feature space rather than optimisation for narrow hard-wire tasks such as multiclass classification. Cross-modal supervision addresses this by aligning representations from different modalities within a shared embedding space \cite{liu2022cross}. In the vision-language setting, text provides supervisory information that is substantially richer than discrete class labels, as it can encode appearance, context, pathology semantics, and relations between concepts. Models trained through image-text alignment can therefore learn embeddings that are not only discriminative but also reusable across downstream tasks, enabling zero-shot classification, similarity search, and cross-modal retrieval \cite{radford2021learning}.

Although vision-language models have advanced rapidly in general computer vision and several areas of medical imaging, their applicability to wireless capsule endoscopy has not been systematically investigated. Existing medical multimodal models have been developed largely using domains such as radiology, where image-text pairs arise naturally from clinical reports, or from broad biomedical corpora with limited relevance to capsule endoscopy. In contrast, WCE presents a distinct setting characterised by uncontrolled image acquisition, subtle pathology manifestation, and the absence of naturally paired textual supervision at scale. To our knowledge, direct zero-shot transfer of relevant medical-domain vision-language models to WCE has not previously been benchmarked, and no domain-specific image-text alignment framework has yet been introduced for this modality. This gap motivates both a first targeted evaluation of existing models on WCE and the development of a dedicated multimodal representation learning approach for capsule endoscopy.

To address this gap, a WCE-specific CLIP-style vision-language framework, termed CapCLIP, is proposed. In this framework, image-text pairs composed of WCE frames and clinically relevant textual descriptions are aligned within a shared embedding space, allowing supervision to move beyond fixed class labels. Because naturally paired image-text data are not available for WCE, textual descriptions are derived from frame-level annotations using pathology-aware caption templates informed by standardised nomenclature and lesion semantics. Through this design, structured semantic information is incorporated into the training of both visual and textual representations, yielding multimodal embeddings suitable for downstream tasks such as zero-shot classification and cross-modal retrieval.

The proposed framework is evaluated in a zero-shot setting on previously unseen, out-of-distribution WCE datasets and is compared with relevant open-source vision and vision-language foundation models. Three downstream tasks are considered: K-nearest neighbour classification as a probe of visual embedding quality, CLIP-style image-text classification as a measure of cross-modal discriminative capability, and text-to-image retrieval as an assessment of language-guided querying for WCE analysis. This evaluation framework enables both the first targeted benchmarking of existing multimodal models on WCE and the assessment of whether domain-specific image-text alignment offers measurable advantages for the modality. The findings indicate that language-guided representation learning can yield embeddings that are more transferable and semantically informative for capsule endoscopy than those produced by existing baseline models.

The main contributions of this paper are twofold. First, a systematic zero-shot benchmarking framework is developed to evaluate the transferability of relevant open-source vision and vision-language foundation models on wireless capsule endoscopy data. Second, a domain-specific multimodal vision-language framework, termed CapCLIP, is proposed for WCE analysis, enabling language-grounded representation learning and cross-modal alignment for downstream classification and retrieval tasks.

The remainder of the paper is structured as follows. Related work is reviewed in Section II, the proposed method is presented in Section III, the experimental setup is described in Section IV, the results are reported in Section V, the implications of the results are discussed in Section VI, and the paper is concluded in Section VII.

\section{Related Work}
\label{sec:rel-work}
\subsection{Machine Learning for WCE}
Recent deep learning research in wireless capsule endoscopy has focused primarily on frame-level classification and lesion detection, often using convolutional neural networks (CNNs), vision transformers (ViTs), or task-specific architectural modifications. Recent review studies have outlined the evolution of machine learning in WCE, showing a transition from early approaches based on handcrafted feature extraction to end-to-end learning frameworks built on neural networks \cite{elgammal2025survey, karar_handcraft, li2012tumor}.

Despite the large body of work on ML-assisted WCE analysis, most reported methods remain vision-only \cite{Wahab_hollistic}. Moreover, strong performance is commonly reported under retrospective evaluation protocols in which training and test splits are derived from the same dataset. Only a limited number of studies have assessed performance on heterogeneous test sets, and these have generally reported noticeable degradation in performance \cite{wahab2024federatedwce}. Such loss of performance under distribution shift has also been identified as a major research gap that limits the clinical usefulness of current ML-based WCE systems.

Overall, recent WCE computer-aided diagnosis (CAD) research has improved task-specific and pathology-specific performance, particularly under within-dataset evaluation settings \cite{wahab2023}. However, the literature remains dominated by vision-only approaches, and the use of cross-modal supervision to improve semantic richness, transferability, and zero-shot generalisation in WCE has not been systematically explored.

\subsection{Vision-Language Representation Learning and Multimodal Foundation Models}
Conventionally, neural networks are most often trained using supervision from a single modality, typically through fixed categorical labels. Although effective for narrowly defined prediction tasks, such supervision provides limited semantic structure and may not support robust generalisation under real-world variability. Cross-modal supervision offers an alternative by aligning information from different modalities within a shared embedding space. In the vision-language setting, this is particularly appealing because text can provide richer supervisory information than mere class labels, including descriptive, contextual, and relational semantics \cite{krishna2017visual}. CLIP-style learning has become a prominent example of this paradigm, showing that aligned image-text representations can support zero-shot classification and cross-modal retrieval \cite{radford2021learning}.

Recent years have also seen the emergence of foundation models in medical imaging and endoscopy. In the endoscopic domain, large-scale representation learning has begun to appear through vision-only foundation models \cite{endossl, endofm, endofmlv}, but a multimodal foundation model has not yet been established for WCE. In the broader medical imaging literature, multimodal vision-language models have developed more rapidly. Early approaches such as ConVIRT \cite{ConVIRT} and GLoRIA \cite{2021gloria} demonstrated the value of image-text alignment for transferable medical representations, followed by models such as MedCLIP \cite{wang2022medclip} and BioMedCLIP \cite{zhang2023biomedclip} that extended contrastive vision-language learning to larger and more specialised medical corpora. More recent systems, including Med-Flamingo \cite{med-flamingo}, SkinGPT \cite{zhou2023skingpt}, PathChat \cite{Pathchat}, EyeCLIP \cite{shi2024eyeclip}, and MedImageInsight \cite{codella2024medimageinsight}, have further expanded multimodal learning across diverse imaging domains and downstream tasks. However, most of these models have been developed in areas such as radiology, dermatology, pathology, and ophthalmology, where image-text pairing is more naturally available.

WCE remains comparatively underexplored in this regard. Its images are acquired under uncontrolled conditions, pathological findings can be subtle and heterogeneous, and naturally paired textual descriptions are not available at scale. Consequently, the transferability of existing medical vision-language models to WCE remains uncertain, while no domain-specific multimodal foundation model has yet been established for the modality. This leaves a clear research gap between recent advances in multimodal medical representation learning and the needs of WCE analysis, particularly with respect to semantic representation. 

\begin{figure} 
    \centering
    \includegraphics[scale=0.75]{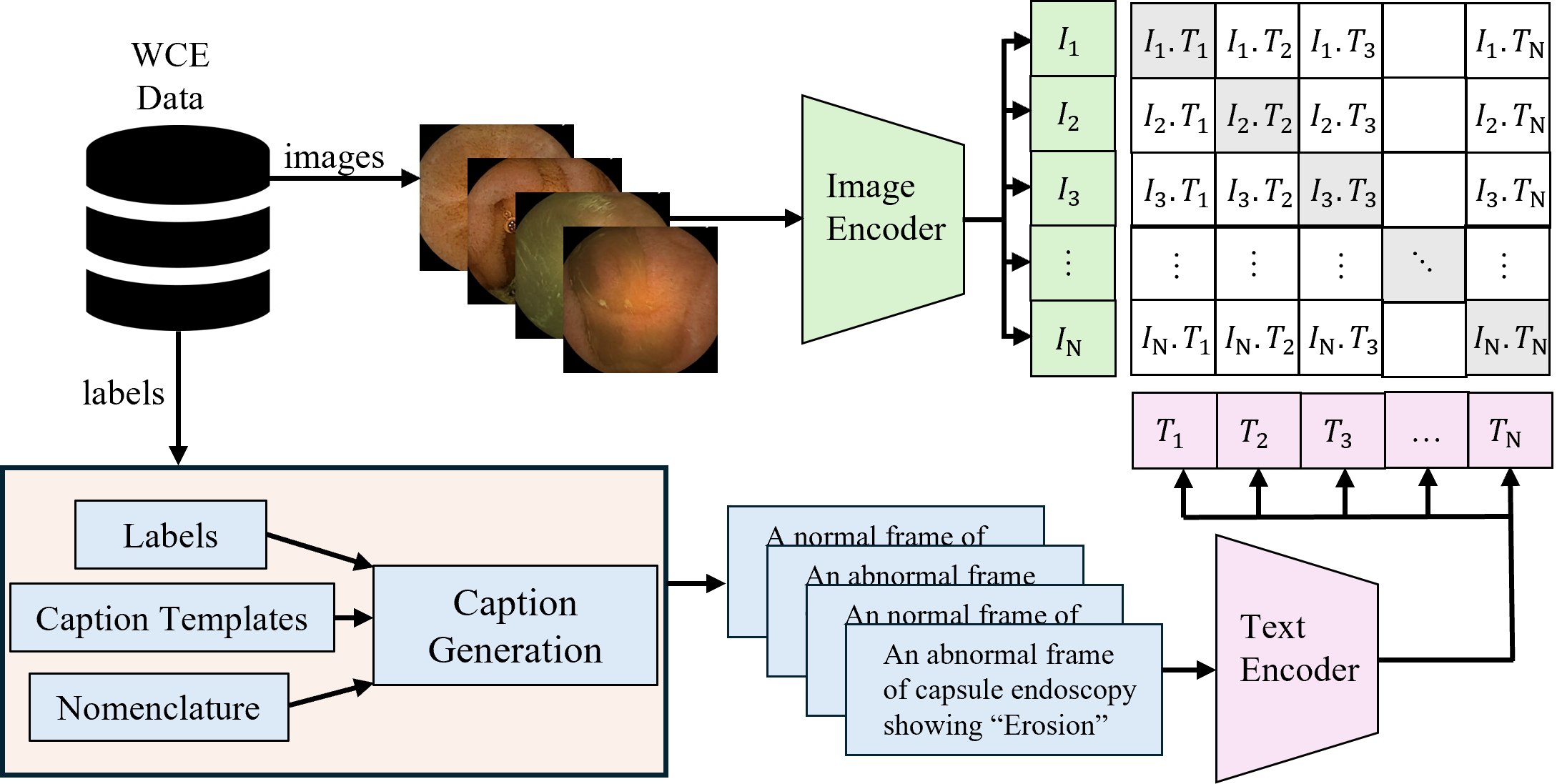}
    \caption{Training Framework for CapCLIP. Text captions generation (in orange) using labels, caption templates, and nomenclature; Image embeddings generated by image encoder (in green) and text embeddings generated by text encoder (in purple) are aligned using symmetric cross entropy loss. The grey squares in the similarity matrix represent the cross-modal labels.}
    \label{fig:meth1}
\end{figure}

\section{Proposed Method}
\label{sec:prop-meth}
In this section, the methodology adopted to formulate representation learning for WCE analysis as vision-language cross-modal alignment is presented in detail. First, the overall framework for training CapCLIP is described. Next, the approach used to transform frame labels into textual descriptions is presented. Finally, the CapCLIP variants and implementation details used for training are described.

\subsection{Framework for Training CapCLIP}
An overview of the CapCLIP training framework is shown in Figure \ref{fig:meth1}. The training process begins with WCE images, and their corresponding labels obtained from the datasets. These labels, which initially function only as class identifiers, are passed to the caption generation block, shown in orange. This block uses predefined caption templates together with domain-specific nomenclature to convert raw labels into textual descriptions that encode both visual characteristics and pathological relevance. Further details of the caption generation process are provided in the following subsection.

The generated captions are then paired with images according to their labels. All images belonging to the same class are assigned one of the captions generated for that class. To maintain representational balance, captions are distributed uniformly within each class, such that each caption in the pool is assigned evenly across the images of that class. Once these image-text pairs are formed, they are passed through the CapCLIP training pipeline. Mini-batches of images are provided to the image encoder, while the corresponding captions are fed to the text encoder, producing visual and textual embeddings, respectively.

For the image encoder, a Vision Transformer (ViT) \cite{ViT} is used, and two variants are evaluated: ViT-B/16 and ViT-L/14, which differ in model size and parameter count. ViT is selected because of its strong performance on image recognition and vision-related multi-task learning benchmarks. The image encoder is initialised with OpenAI CLIP weights, since training from scratch would require data scale and computational resources that are not feasible in the present setting.
The text encoder follows the original CLIP framework \cite{radford2021learning}. It consists of a 12-layer Transformer with 8 attention heads and a text embedding dimension of 512. For tokenisation, a lowercased byte-pair encoding (BPE) scheme is used with a vocabulary of 49,512 tokens \cite{sennrich2016neural_BPE}. The maximum context length, corresponding to the number of tokens per caption, is fixed at 76, consistent with the OpenAI CLIP training setup.
Following encoding, visual and textual representations are L2-normalised so that both modalities lie in a common hyperspherical embedding space. Cross-modal similarity is then evaluated by computing cosine similarity between every image embedding and every text embedding within the mini-batch. This produces an image-text similarity matrix that forms the basis for contrastive optimisation.
Training is driven by a symmetric cross-entropy objective applied in both image-to-text and text-to-image directions. Under this objective, matched pairs are encouraged to attain high similarity, whereas unmatched pairs within the batch are pushed apart. In this way, the visual and textual encoders are jointly optimised to learn aligned multimodal representations. The mathematical form of this objective is given below.

Let \(f_{\theta}(x)\) denote the image encoder and \(g_{\phi}(t)\) the text encoder, where an input image \(x\) and a text input \(t\) are mapped to 512-dimensional embedding vectors. For the \(i\)th paired sample in a mini-batch, the \(\ell_2\)-normalised visual and textual embeddings are denoted by \(u_i\) and \(v_i\), respectively:

\begin{equation}
u_i=\frac{f_{\theta}(x_i)}{\|f_{\theta}(x_i)\|}, \qquad
v_i=\frac{g_{\phi}(t_i)}{\|g_{\phi}(t_i)\|}
\end{equation}

For a mini-batch of size \(N\), cross-modal similarity is computed between every visual and textual embedding, yielding a similarity matrix \(S \in \mathbb{R}^{N \times N}\) defined as:

\begin{equation}
S_{ij}=\frac{u_i \cdot v_j}{\tau}
\end{equation}

where \(\tau\) is a learnable temperature parameter. Based on this similarity matrix, the symmetric cross-entropy loss is applied in two directions.

From image to text, the supervision is provided by the text modality:
\begin{equation}
L_{i \rightarrow t}
=
-\frac{1}{N}
\sum_{i=1}^{N}
\log
\left(
\frac{\exp(S_{ii})}
{\sum_{j=1}^{N}\exp(S_{ij})}
\right)
\end{equation}

From text to image, the supervision is provided by the image modality:
\begin{equation}
L_{t \rightarrow i}
=
-\frac{1}{N}
\sum_{i=1}^{N}
\log
\left(
\frac{\exp(S_{ii})}
{\sum_{j=1}^{N}\exp(S_{ji})}
\right)
\end{equation}

The final CLIP objective is obtained by averaging the two directional losses:
\begin{equation}
L_{\text{CLIP}}
=
\frac{1}{2}
\left(
L_{i \rightarrow t}
+
L_{t \rightarrow i}
\right)
\end{equation}

\begin{figure} 
    \centering
    \includegraphics[scale=0.75]{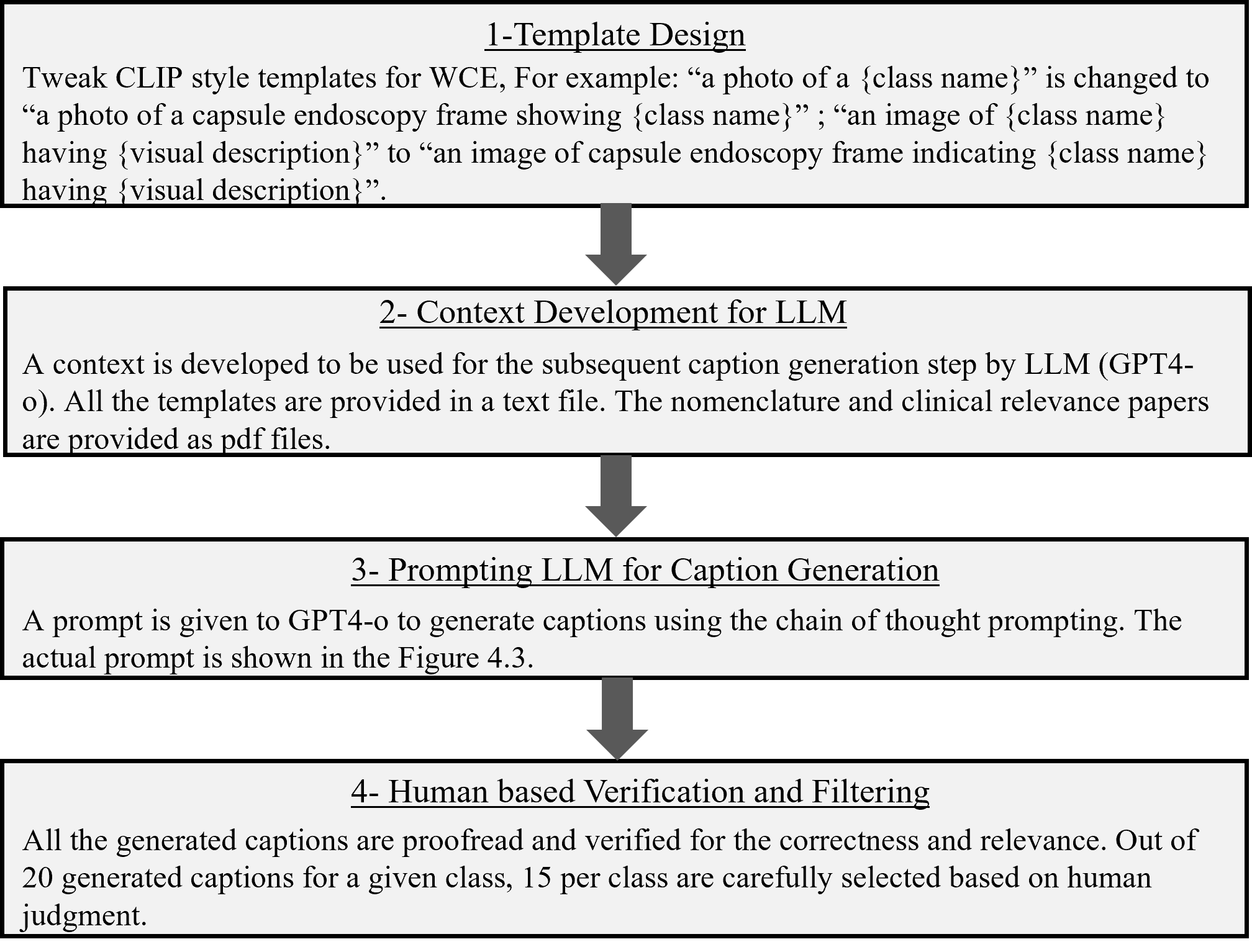}
    \caption{Caption generation stepwise process.}
    \label{fig:meth2}
\end{figure}

\subsection{From Labels to Captions}
Captions are generated by combining frame labels, caption templates, and domain knowledge. The overall caption generation pipeline is illustrated in Figure \ref{fig:meth2} and consists of four steps. A caption template first provides the linguistic structure of the sentence without including any class-specific content. A small number of such templates have previously been used in the literature for CLIP-style models \cite{meta_prompting}, and these are adopted here with slight modifications for the WCE domain. Class-specific information is then inserted into the templates to form complete captions. This information may include the class name, visual description, alternative terminology, disease relevance, sub-pathology types within the taxonomy, and parent pathology category. Much of this content is derived from Delphi-consensus studies on the nomenclature, visual characteristics, and clinical relevance of frame-level WCE findings \cite{leenhardt2019vascular, leenhardt2020ulcerative}.

LLMs are well suited to generating captions when guided by human-written prompts and supported by domain-specific textual context \cite{tan2024large}. After the caption templates are finalised in the first step, a contextual knowledge base is assembled in the second step from the detailed literature presented in \cite{wahab2023}. The relevant papers and reports are supplied in PDF format, while the templates are provided as text inputs. In the third step, captions are generated by prompting the LLM with this domain context using the prompt formulation provided in Appendix \ref{app:caption_details}, Figure \ref{fig:appendix_prompt}. All generated captions are then proofread and verified for correctness. In the final step, fifteen captions judged to be the most relevant to the intended context are selected for use.

Examples of the resulting captions for several classes are provided in Appendix \ref{app:caption_details}, Table \ref{tab:appendix_captions}. The captions are designed not only to describe how a finding appears visually, but also to encode its position within the pathology taxonomy and its clinical relevance. For example, the caption for Erythema does not remain limited to surface appearance but also places the finding under its parent lesion category by referring to it as a vascular lesion. Similarly, clinical relevance derived from the literature is incorporated into the generated captions for Abnormal, including its association with Crohn’s disease. In this way, the caption generation process injects structured semantic information into image-text pairing beyond simple label naming.

\subsection{CapCLIP Variants}
\label{sec:capclip_variants}
Three CapCLIP modes are trained to examine differences in downstream performance. The first mode, CapCLIP-B, is trained by pairing captions with binary classes from the training data, namely Normal, representing all normal frames, and Abnormal, representing all pathological frames. The second mode, CapCLIP-M, is trained in a multiclass setting, where captions are paired with multiple distinct class labels from the training datasets.

The third mode adopts a hybrid formulation that combines both binary and multiclass pairings. In this setting, each image is associated with both a binary caption and a multiclass caption, which doubles the number of training samples relative to the other two modes. This hybrid model is referred to as CapCLIP-MIX.

\subsection{Training Details}
\label{sec:train_det}
The experiments are conducted on a machine equipped with an NVIDIA A100 GPU with 40 GB memory. All training runs are performed for 30 epochs, with each experiment taking approximately 12 hours. In practice, overfitting is consistently observed between epochs 18 and 20, while the best validation performance is typically obtained between epochs 15 and 20. Optimisation is carried out using Adam together with a cosine learning rate scheduler. The learnable temperature parameter
$\tau$ is initialised to 0.07, following the original CLIP implementation.

The maximum batch size used in the experiments is 128. After each epoch, the CLIP loss is evaluated on the validation set, and a model checkpoint is saved whenever the validation loss improves. For each configuration, defined by the CapCLIP mode and image encoder, three models are trained using three different train/validation splits, resulting in three independent checkpoints per configuration. During evaluation, all three checkpoints are loaded, and the reported metrics are averaged across these runs.

\section{Experimental Setup}
This section describes the datasets, baseline models, evaluation tasks, and performance metrics used to assess CapCLIP and the compared foundation models on wireless capsule endoscopy data. The evaluation is designed to examine zero-shot transferability under out-of-distribution conditions, with all test datasets kept entirely unseen during training. Three downstream tasks are considered, namely K-nearest neighbour classification, CLIP-style image-text classification, and text-to-image retrieval. Through this setup, both the transferability of existing foundation models to WCE and the effectiveness of the proposed domain-specific alignment strategy are evaluated.
\subsection{Datasets}
Five WCE datasets are used in this study, with each dataset assigned to one of three roles: training, validation, or testing. The numbers of images, class composition, and dataset roles are summarised in Table \ref{tab:datasets}. In contrast to many previous studies, where training, validation, and testing are all derived from the same dataset, the present work adopts a strict out-of-distribution evaluation protocol. Under this protocol, all test datasets remain entirely unseen during both training and validation.

For training and validation, the SEE-AI and Kvasir-Capsule datasets are selected based on two considerations. First, when combined, they provide a sufficiently large number of images to support training of vision-language models. Second, together they cover a diverse range of gastrointestinal findings, thereby offering a broader supervisory signal for representation learning. The test phase is conducted on previously unseen datasets in order to assess cross-dataset transferability under out-of-distribution conditions.

The total number of annotated images available in the original datasets may exceed the number used in the present experiments. Since most datasets are affected by class imbalance, with normal and normal-variant categories dominating the distribution, these majority classes are under-sampled to reduce imbalance in the learned representation space. This under-sampling is applied to Kvasir-Capsule and Galar. The class distributions corresponding to the training and test splits are provided in the Appendix \ref{sec:appendix_class_dist}.

\begin{table*}[t]
\centering
\caption{Characteristics of datasets involved in the experiments across training, validation, and testing phases.}
\label{tab:datasets}
\renewcommand{\arraystretch}{1.15}
\begin{tabular}{p{2.2cm} p{2.3cm} p{9.2cm} p{2.2cm}}
\hline
\textbf{Dataset} & \textbf{No. of Images/Frames} & \textbf{No. of Classes; Names} & \textbf{Split} \\
\hline
SEE-AI & 16914 & 12; Angiodysplasia, Bleeding, Diverticulum, Erosion, Erythema, Lymphangiectasia, Lymph follicle, Normal, Polyp, SMT, Stenosis, Vein & Train and Validation \\

Kvasir Capsule & 6982 & 10; Angiectasia, Blood, Erosion, Erythema, Foreign body, Lymphangiectasia, Normal, Polyp, Reduced Mucosal View, Ulcer & Train and Validation \\

CrohnIPI & 3484 & 7; Aphthous, Erythema, Normal, Edema, Stenosis, Superficial Ulcer, Deep Ulcer & Test Only \\

KID2 & 1302 & 4; Inflammatory, Normal, Polypoids, Vascular & Test Only \\

Galar & 3705 & 12; Active Bleeding, Angiectasia, Blood, Cancer, Erosion, Erythema, Hematin, IBD, Lymphangiectasia, Normal, Polyp, Ulcer & Test Only \\
\hline
\end{tabular}
\end{table*}

\subsection{Baseline Foundation Models}
\label{baselines}

Baseline models were selected to ensure relevance and comparability in the benchmarking study. Inclusion was restricted to models that followed a foundation-model pretraining paradigm, had been pretrained on either endoscopic data or broad medical image-text corpora with potential relevance to WCE, demonstrated strong transferability in prior work, and were supported by publicly available code or pretrained weights. Generative multimodal models developed for medical dialogue or report generation were excluded, as the present study focuses on representation alignment rather than language generation. One exception was made for CLIP \cite{radford2021learning}, which, although not trained on medical or endoscopic data, was included because of its strong robustness under distribution shift and its widespread use as a competitive vision-language baseline.

\textbf{EndoSSL.}
EndoSSL \cite{endossl} is a vision-only foundation model for endoscopy that pretrains Vision Transformers using Masked Siamese Networks (MSN). Its pretraining strategy is based on self-supervised alignment between augmented image views and is designed to learn transferable endoscopic representations from large-scale unlabeled colonoscopy and laparoscopy data. Although not multimodal, it is included because it represents one of the few dedicated foundation models available in the endoscopic domain.

\textbf{EndoFM.}
EndoFM \cite{endofm} is a vision-only endoscopic foundation model built on a video-transformer backbone with space-time attention and dynamic spatial-temporal positional encoding. Pretraining is conducted in a teacher-student setting using objectives designed to match global and local video views and to learn invariances across motion and temporal scale. The model is trained on a large endoscopy corpus spanning colonoscopy, gastroscopy, and laparoscopy, and is included here as a representative large-scale vision-only endoscopic baseline.

\textbf{EndoFM-LV.}
EndoFM-LV \cite{endofmlv} extends EndoFM from short clips to minute-long endoscopic sequences. It retains the teacher-student self-supervised formulation but introduces masked token matching to capture finer spatial and temporal structure over longer streams. Since it reports improved performance over both EndoSSL and EndoFM in its original evaluation, it is included as an additional endoscopic foundation baseline.

\textbf{CLIP.}
CLIP \cite{radford2021learning} is a general-domain vision-language model trained on 400 million web-collected image-text pairs using contrastive image-text alignment. The model jointly trains image and text encoders with a symmetric cross-entropy objective and has shown strong zero-shot transfer across a wide range of computer vision benchmarks. Although its pretraining data are neither medical nor endoscopic, it is included as an exception because of its demonstrated robustness, strong representation quality, and role as a widely adopted baseline in multimodal evaluation studies.

\textbf{MedCLIP.}
MedCLIP \cite{wang2022medclip} adapts CLIP-style learning to the medical domain while addressing two issues specific to medical image-text training: limited paired data and false negatives within contrastive batches. To address these, it uses combinatorial pair construction and a knowledge-guided semantic matching loss instead of standard InfoNCE. Despite being presented as a general medical VLM, its training data are derived entirely from radiology corpora, with a Swin Transformer image encoder and a BioClinicalBERT-based text encoder.

\textbf{BioMedCLIP.}
BioMedCLIP \cite{zhang2023biomedclip} is a general biomedical vision-language model pretrained on PMC-15M, a large corpus of figure-caption pairs extracted from PubMed Central articles. In contrast to CLIP, it employs PubMedBERT as the text encoder to improve alignment with biomedical language. Its pretraining data span multiple medical fields, including radiology, histopathology, dermatology, ultrasound, and endoscopy, making it a strong candidate for testing cross-domain multimodal transfer to WCE.

\textbf{MedImageInsight.}
MedImageInsight \cite{codella2024medimageinsight} is a large-scale two-tower medical image-text embedding model developed to support diverse downstream tasks across multiple medical imaging domains. It uses DaViT as the image encoder, a large transformer-based text encoder, and a unified contrastive objective that combines image-text and image-label supervision. Because it is trained across a broad set of medical imaging modalities, including endoscopy, and supports zero-shot classification and retrieval tasks, it serves as one of the most relevant multimodal baselines for comparison.

\begin{figure} 
    \centering
    \includegraphics[scale=0.6]{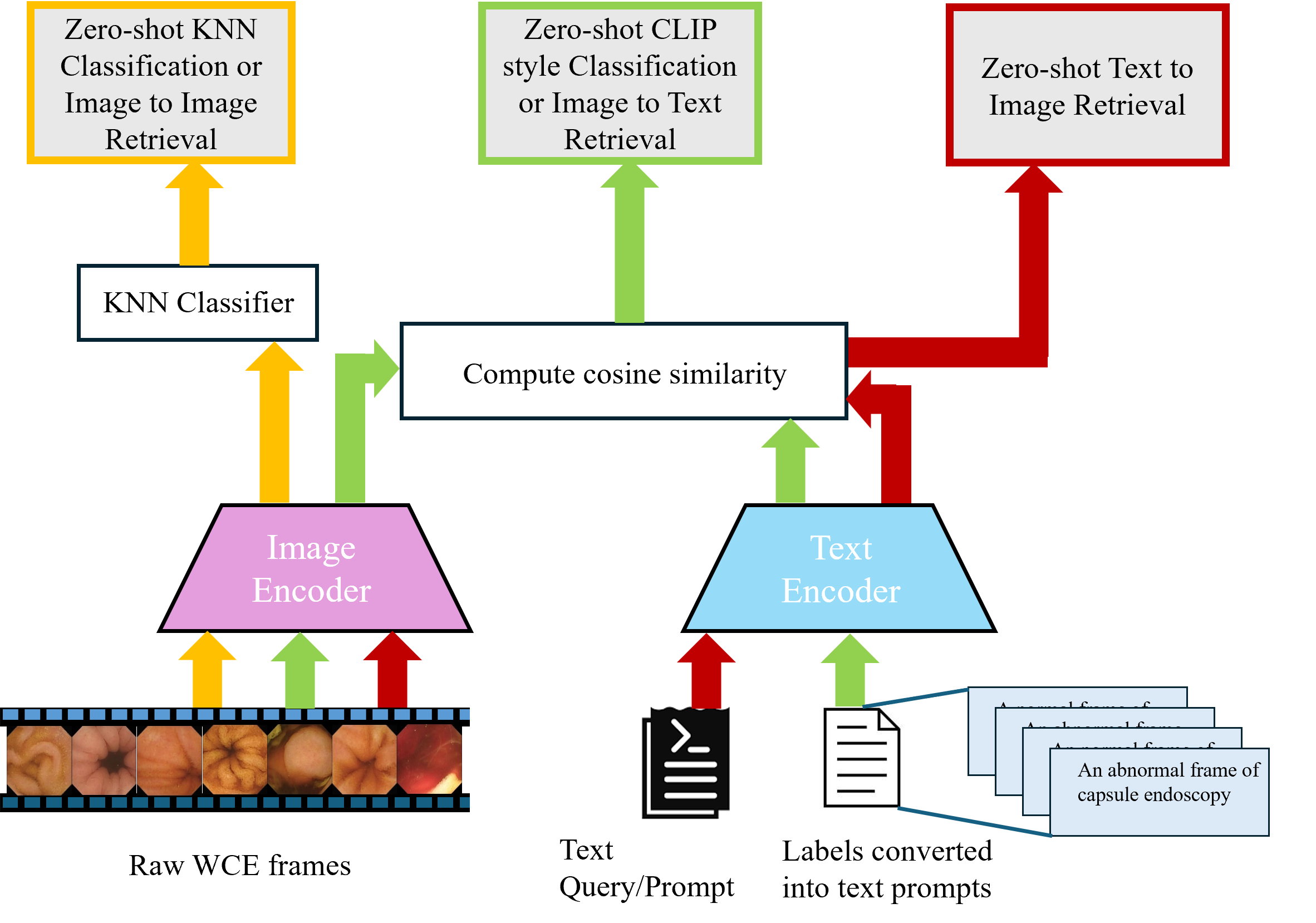}
    \caption{Overview of the evaluation framework for three downstream tasks: 1) Zero-shot KNN classification or Image to Image Retrieval (flow path in yellow), 2) Zero-shot CLIP style Classification or Image to Text Retrieval (flow path in green), 3) Zero-shot Text to Image Retrieval (flow path in red).}
    \label{fig:eval}
\end{figure}

\subsection{Evaluation Tasks}
All baseline foundation models included in the benchmarking study, together with CapCLIP, are evaluated under a zero-shot setting across three downstream tasks, as illustrated in Figure \ref{fig:eval}. In the present work, zero-shot evaluation is used to examine the transferability of learned representations to datasets that remain entirely unseen during training. Although these evaluation sets belong to the same broad application domain, namely wireless capsule endoscopy, they differ from the training data in clinically and technically relevant ways. Two forms of distribution shift are particularly important in this setting. The first is semantic shift, where the evaluation data contain pathology distributions or lesion categories not represented in the same way during training. The second is feature drift, where the data originate from different hospitals, device manufacturers, or acquisition protocols, resulting in variation in image appearance and dataset characteristics. Together, these conditions provide a practical setting for assessing zero-shot transferability under out-of-distribution WCE evaluation.

\textbf{Zero-shot KNN classification.} To assess the quality of the visual embeddings, a K-nearest neighbour (KNN) classification probe is adopted, as shown in Figure \ref{fig:eval}. KNN probing is widely used in representation learning as a direct test of semantic structure in the learned embedding space \cite{codella2024medimageinsight, simCLR, dino}. Since the goal is to evaluate performance on truly unseen datasets, each test image embedding is classified using its k nearest neighbours drawn from the remaining test set, thereby avoiding any dependence on a training-derived support set. This setup evaluates the intrinsic clustering quality of the learned representations with respect to the test-set classes.

For KNN-based classification, k=5 is used. A relatively small value of k provides a more stringent test of clustering behaviour, since fewer neighbours are available to compensate for local misalignment in the representation space. In addition, an odd value reduces the likelihood of tied votes in binary settings. Cosine similarity is adopted as the distance measure, as it is commonly used in vision-language and self-supervised representation learning, particularly when embeddings are L2-normalised. This choice enables similarity to be evaluated primarily through embedding direction rather than magnitude, thereby emphasizing semantic alignment in the learned feature space.

\textbf{Zero-shot CLIP-style image-text classification.} Vision-language models such as CLIP provide a natural mechanism for zero-shot classification because class concepts can be represented directly through text. Whereas vision-only models are commonly evaluated through probes such as linear classification or K-nearest neighbour analysis, VLMs can instead be assessed by converting target class labels into textual prompts and comparing them with image embeddings in the shared latent space. In this setting, the prompts are encoded by the text encoder and the test images by the vision encoder, after which cross-modal similarity scores are computed and normalised with a SoftMax function to obtain class probabilities, as illustrated in Figure \ref{fig:eval}. This protocol has become a standard approach for evaluating vision-language models on downstream classification tasks \cite{qian2024online}.

The same evaluation can also be viewed from a retrieval perspective. Rather than treating the task only as classification, the procedure may be interpreted as image-to-text retrieval, in which a query image is used to identify the most semantically compatible class prompt. In the context of WCE, this formulation is relevant because a given frame, or a set of frames, may be mapped to contextually appropriate textual descriptions drawn from a predefined vocabulary of findings or diagnostic concepts.

To enable this zero-shot setting, the ground-truth labels of each test dataset are converted into textual prompts, with one prompt corresponding to each class. Five prompt sets are defined for every dataset and for each classification mode, namely binary and multiclass, following CLIP-style templates with minor variations to assess prompt sensitivity. Metrics are computed separately for each prompt set, and both the mean performance and the associated variation are reported. Because this task requires a joint image-text embedding space, it is applicable only to vision-language models. Consequently, vision-only endoscopic baselines such as Endo-FM, EndoFM-LV, and Endo-SSL are excluded from this part of the evaluation.

\textbf{Zero-shot text-to-image retrieval.} Zero-shot text-to-image retrieval is also supported naturally by vision-language models, since the shared embedding space allows a query in one modality to retrieve semantically related instances from the other. In contrast to the previous task, where an image is matched against a set of class prompts, text-to-image retrieval considers the reverse direction: a text query is supplied and the model retrieves the most relevant frames from the video. This setting is particularly pertinent to WCE analysis, where a clinician may wish to search for visual evidence of a specific pathology, such as ulcer, erosion, or angioectasia, and retrieve the most representative frames from a long video sequence. In this way, textual queries can serve as targeted retrieval tools for rapid frame prioritisation, video summarisation, and computer-assisted review.

In the present study, zero-shot text-to-image retrieval is implemented by encoding text queries through the text encoder and images through the vision encoder, following the evaluation flow illustrated in Figure \ref{fig:eval}. Both modalities are projected into the shared multimodal embedding space, where similarity is computed using cosine similarity. Images are then ranked according to their similarity to the query, and the highest-ranked results are returned as the retrieval output. As with CLIP-style image-text classification, this task is applicable only to models that learn joint image-text representations.

\subsection{Evaluation Metrics}

Different metrics are used depending on the nature of the downstream task. For classification-oriented evaluations, precision, recall, and F1-score are used as the primary metrics, while binary image-text classification is additionally assessed using threshold-independent measures. For retrieval, ranking-based metrics are adopted to reflect the relevance and ordering of returned results.

\textbf{Classification metrics.}
To assess the classification performance of the learned representations, precision, recall, and F1-score are computed. These metrics are preferred over accuracy, particularly in the presence of class imbalance and multiclass settings, since they provide a more informative account of false positives and false negatives. Precision, recall, and F1-score are defined as
\begin{equation}
\text{Precision}=\frac{TP}{TP+FP}
\end{equation}
\begin{equation}
\text{Recall}=\frac{TP}{TP+FN}
\end{equation}
\begin{equation}
F1=\frac{2 \cdot (\text{Precision} \cdot \text{Recall})}{\text{Precision}+\text{Recall}}
\end{equation}
where \(TP\), \(FP\), and \(FN\) denote true positives, false positives, and false negatives, respectively. These metrics are reported as macro and weighted averages. Macro-averaged scores treat all classes equally, whereas weighted averages account for class support and therefore reflect the underlying class distribution. F1-score is used as the main comparative metric for multiclass evaluation.

\textbf{Binary image-text classification metrics.}
For binary CLIP-style image-text classification, prediction probabilities are available after applying the SoftMax function. Therefore, threshold-independent metrics are used to assess discriminative performance, namely the Receiver Operating Characteristic (ROC) curve and the Precision-Recall curve (PRC), together with their respective areas under the curve, AUROC and AUPRC. The ROC curve plots the true positive rate against the false positive rate across decision thresholds, where
\begin{equation}
TPR=\frac{TP}{TP+FN}, \qquad FPR=\frac{FP}{FP+TN}
\end{equation}
and \(TN\) denotes true negatives. AUROC summarises overall class separability across thresholds, while AUPRC is particularly informative when the positive class is relatively rare. For multiclass CLIP-style classification, macro and weighted precision, recall, and F1-score are used instead, as these are more interpretable than one-vs-rest or pairwise ROC-based extensions in the current setting.

\textbf{Retrieval metrics.}
Text-to-image retrieval performance is evaluated using mean Average Precision (mAP), Recall@\(K\), and Precision@\(K\). Average Precision (AP) measures the ranking quality of retrieved items for a given query and is defined as
\begin{equation}
AP=\frac{1}{R}\sum_{k=1}^{N} P(k)\,\mathrm{rel}(k)
\end{equation}
where \(R\) is the total number of relevant images for a query, \(N\) is the total number of retrieved images, \(P(k)\) denotes the precision at rank \(k\), and \(\mathrm{rel}(k)\) is an indicator function equal to 1 if the image at rank \(k\) is relevant and 0 otherwise. The mean of AP over all queries gives the mAP score.

In standard CLIP-style retrieval, Recall@\(K\) is often treated as a binary measure indicating whether the paired instance is retrieved within the top-\(K\) results. In the present WCE setting, however, a query may correspond to a broader pathology category with multiple relevant frames. A set-based definition of Recall@\(K\) is therefore adopted:
\begin{equation}
\mathrm{Recall}@K=\frac{\text{No. of relevant images in top-}K}{\text{No. of all relevant images}}
\end{equation}
This formulation is more appropriate for clinically meaningful retrieval, where multiple frames may serve as valid matches to the same query. Precision@\(K\) is defined as
\begin{equation}
\mathrm{Precision}@K=\frac{\text{No. of relevant images in top-}K}{K}
\end{equation}
and measures the concentration of relevant results among the top-ranked retrieved images. Together, mAP, Recall@\(K\), and Precision@\(K\) provide a complementary view of retrieval performance, capturing global ranking quality, early recall, and top-rank relevance concentration.

\section{Results}
This section presents the results of the proposed evaluation protocol and compares CapCLIP with the selected baseline foundation models across the three downstream tasks.

\begin{table*}[t]
\centering
\caption{Summary of zero-shot KNN classification performance in terms of weighted F1 across all datasets and both classification modes. The row ``Best CapCLIP'' reports the highest-performing CapCLIP result for each setting. Variant notation: B = CapCLIP-B, M = CapCLIP-M, MX = CapCLIP-MIX, B16 = ViT-B/16, and L14 = ViT-L/14. Detailed per-dataset KNN results are provided in Appendix C.}
\label{tab:knn_summary}
\renewcommand{\arraystretch}{1.08}
\setlength{\tabcolsep}{8pt}
\footnotesize
\begin{tabular}{lcccccc}
\hline
\textbf{Model} & \multicolumn{2}{c}{\textbf{KID2}} & \multicolumn{2}{c}{\textbf{CrohnIPI}} & \multicolumn{2}{c}{\textbf{Galar}} \\
\cline{2-3} \cline{4-5} \cline{6-7}
 & \textbf{Bin} & \textbf{Mul} & \textbf{Bin} & \textbf{Mul} & \textbf{Bin} & \textbf{Mul} \\
\hline
EndoFM          & 0.793 & 0.745 & \textbf{0.929} & 0.872 & 0.958 & \textbf{0.808} \\
EndoFM-LV       & 0.752 & 0.674 & 0.899 & 0.826 & 0.950 & 0.773 \\
EndoSSL         & 0.781 & \textbf{0.753} & 0.928 & \textbf{0.876} & 0.960 & 0.804 \\
CLIP-OpenAI     & \textbf{0.806} & 0.737 & 0.904 & 0.838 & 0.958 & 0.790 \\
MedCLIP         & 0.731 & 0.639 & 0.870 & 0.792 & 0.918 & 0.735 \\
BioMedCLIP      & 0.798 & 0.713 & 0.917 & 0.849 & 0.950 & 0.785 \\
MedImageInsight & 0.783 & 0.714 & 0.927 & 0.875 & \textbf{0.964} & 0.806 \\
\hline
Best CapCLIP    & \textbf{0.878} & \textbf{0.837} & \textbf{0.968} & \textbf{0.919} & \textbf{0.967} & 0.806 \\
Variant         & B-B16 & M-B16 / M-L14 & M-L14 & M-L14 & M-B16 / M-L14 & M-L14 \\
Gain (\%)       & +8.9 & +11.2 & +4.2 & +4.9 & +0.3 & -0.2 \\
\hline
\end{tabular}
\end{table*}

\begin{figure} 
    \centering
    \includegraphics[scale=0.4]{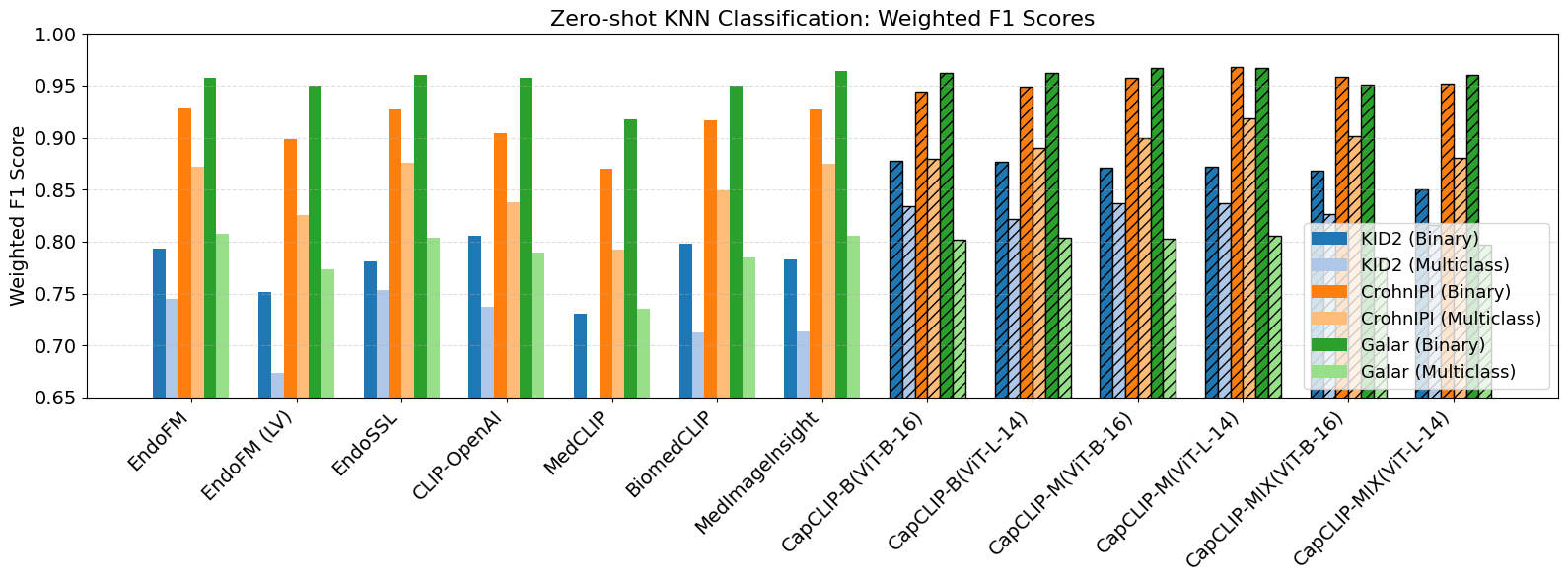}
    \caption{Zero-shot KNN classification: Comparison of weighted F1 scores for CapCLIP and other baselines for binary and multiclass settings. CapCLIP models are shown in stripped colours for distinction. Other than Galar dataset (green colour), CapCLIP models show better performance than the baselines.}
    \label{fig:knngraphics}
\end{figure}

\begin{figure} 
    \centering
    \includegraphics[scale=0.4]{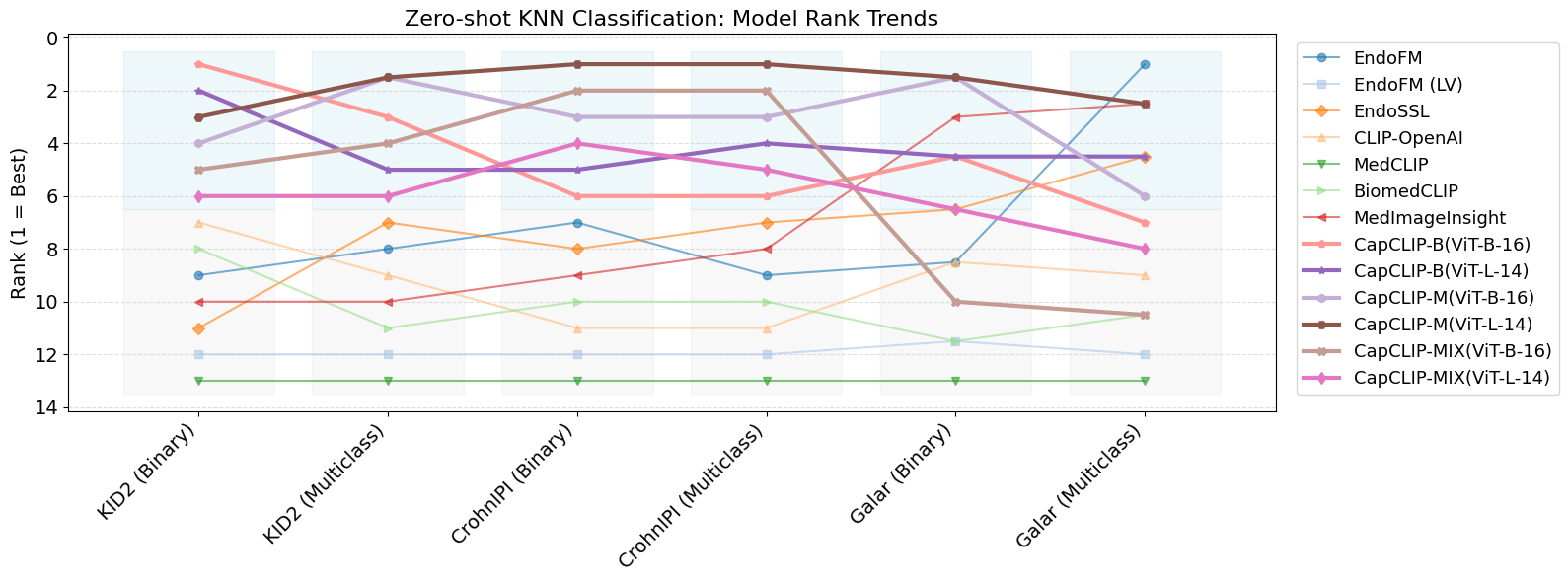}
    \caption{Zero-shot KNN classification: Model rankings based on weighted F1 score across different datasets. CapCLIP models (thicker lines) persistently rank higher (light blue background) across most datasets than the baselines.}
    \label{fig:knn_ranktrends}
\end{figure}
\subsection{Zero-shot KNN Classification}

Zero-shot KNN classification is used as a direct probe of the semantic structure of the learned embedding space. A summary of weighted F1 scores across all datasets and both classification modes is provided in Table~\ref{tab:knn_summary}, while the corresponding visual trends and model-ranking behaviour are shown in Fig.~\ref{fig:knngraphics} and Fig.~\ref{fig:knn_ranktrends}, respectively. Detailed per-dataset binary and multiclass KNN results are reported in Appendix~C.

\textbf{Binary classification.}
In the binary setting, CapCLIP achieves the clearest gains on KID2 and CrohnIPI. On KID2, the best-performing CapCLIP model surpasses the strongest baseline by approximately 9\% in weighted F1, indicating that the learned embeddings generalise well even on a relatively small but carefully annotated dataset. Among the baselines, CLIP-OpenAI provides the strongest binary performance on KID2, outperforming several more domain-specific endoscopic and medical models. This suggests that large-scale language-supervised pretraining can yield strong transferability even without explicit medical-domain specialisation. Nevertheless, all CapCLIP variants remain ahead of the baseline set, with the binary and multiclass CapCLIP modes generally performing better than the hybrid variant.

A similar pattern is observed on CrohnIPI, where all CapCLIP models outperform the baselines. In this case, the best CapCLIP result exceeds the strongest baseline by more than 4\% in weighted F1. Among the baselines, EndoFM, EndoSSL, and MedImageInsight provide the most competitive results, with EndoFM being the strongest baseline in the binary setting. Overall, these results indicate that the domain-specific image-caption alignment used in CapCLIP yields embeddings that transfer more effectively than both general-domain and medical-domain baselines under cross-dataset evaluation.

In contrast, the performance gap on Galar is much smaller. Here, the best CapCLIP result is only marginally above the best baseline, and the overall performance of most models is tightly clustered. This behaviour is indicative of a saturated regime in which KNN classification becomes comparatively less discriminative as an evaluation tool. A likely explanation is the structure of the Galar dataset itself, which is derived from raw WCE videos and contains substantial redundancy, including many visually similar neighbouring frames. Under a leave-one-out KNN setup, such redundancy can make local neighbourhood classification relatively easy even for weaker representations. Importantly, this effect is specific to the probing behaviour of KNN and should not be assumed to transfer directly to the more demanding cross-modal tasks considered later.

\textbf{Multiclass classification.}
In the multiclass setting, the advantage of CapCLIP remains evident on KID2 and CrohnIPI. On KID2, EndoFM and EndoSSL form the strongest baseline group, with EndoSSL slightly ahead in weighted F1. However, the best CapCLIP model improves upon the best baseline by more than 11\%, showing that the learned embeddings preserve stronger class-level structure under finer semantic discrimination. On CrohnIPI, the CapCLIP-M variants again provide the best results, with weighted F1 improvements of up to 4.9\% over the strongest baseline. Among the baselines, EndoFM, EndoSSL, and MedImageInsight remain the most competitive, but none match the best-performing CapCLIP models.

The Galar multiclass results again differ from the other datasets. In this setting, EndoFM slightly outperforms all CapCLIP variants, while EndoSSL, MedImageInsight, and CLIP-OpenAI remain close behind. The performance gap is narrow, and in some cases slightly negative for CapCLIP relative to the best baseline. This is consistent with the earlier observation that the redundancy and local similarity structure of Galar reduce the discriminative value of leave-one-out KNN probing, particularly when many neighbouring frames are visually alike.

The overall KNN trends are illustrated in Fig.~\ref{fig:knngraphics}, where the weighted F1 scores are summarised across all datasets and both classification modes. The separation between CapCLIP and the baselines is visually most pronounced for KID2 and CrohnIPI, whereas the Galar results are more compressed. Complementing this view, Fig.~\ref{fig:knn_ranktrends} shows the ranking trajectories of all models across datasets and modes. CapCLIP variants occupy the higher-ranking positions in most cases, with the CapCLIP-M models showing the most consistent overall behaviour. In contrast, MedCLIP and EndoFM-LV tend to rank lower across settings. Taken together, these results indicate that CapCLIP learns embeddings with stronger zero-shot clustering structure than the compared baselines on two of the three evaluation datasets, while also revealing that KNN probing on highly redundant video-derived datasets may underestimate differences in representation quality.

\begin{table*}[t]
\centering
\caption{Zero-shot CLIP-style binary image-text classification results. Mean AUROC and mean AUPRC are reported across different prompt sets. The best baseline in each column is marked with an asterisk (*), while the best CapCLIP result is shown in bold.}
\label{tab:clip_binary}
\renewcommand{\arraystretch}{1.12}
\setlength{\tabcolsep}{5pt}
\footnotesize
\begin{tabular}{lcccccc}
\hline
\textbf{Model} & \multicolumn{2}{c}{\textbf{KID2}} & \multicolumn{2}{c}{\textbf{CrohnIPI}} & \multicolumn{2}{c}{\textbf{Galar}} \\
\cline{2-3} \cline{4-5} \cline{6-7}
 & \textbf{AUROC} & \textbf{AUPRC} & \textbf{AUROC} & \textbf{AUPRC} & \textbf{AUROC} & \textbf{AUPRC} \\
\hline
CLIP-OpenAI         & 0.618* & 0.545* & 0.709 & 0.621 & 0.583 & 0.784 \\
MedCLIP             & 0.507  & 0.448  & 0.518 & 0.417 & 0.507 & 0.744 \\
BioMedCLIP          & 0.570  & 0.508  & 0.625 & 0.509 & 0.494 & 0.774 \\
MedImageInsight     & 0.612  & 0.545* & 0.781* & 0.692* & 0.600* & 0.813* \\
\hline
CapCLIP-B (ViT-B/16)   & \textbf{0.860} & \textbf{0.842} & 0.960 & 0.942 & 0.893 & 0.958 \\
CapCLIP-B (ViT-L/14)   & 0.852 & 0.820 & \textbf{0.966} & \textbf{0.951} & 0.904 & 0.964 \\
CapCLIP-M (ViT-B/16)   & 0.800 & 0.750 & 0.927 & 0.897 & 0.808 & 0.926 \\
CapCLIP-M (ViT-L/14)   & 0.824 & 0.724 & 0.950 & 0.923 & \textbf{0.921} & \textbf{0.970} \\
CapCLIP-MIX (ViT-B/16) & 0.833 & 0.806 & 0.958 & 0.943 & 0.829 & 0.926 \\
CapCLIP-MIX (ViT-L/14) & 0.689 & 0.615 & 0.916 & 0.890 & 0.807 & 0.924 \\
\hline
Improvement over best baseline (\%) & +39.2 & +54.5 & +23.7 & +37.4 & +53.5 & +19.3 \\
\hline
\end{tabular}
\end{table*}

\begin{figure*}[t]
    \centering
    \begin{subfigure}[t]{0.7\textwidth}
        \centering
        \includegraphics[width=\linewidth]{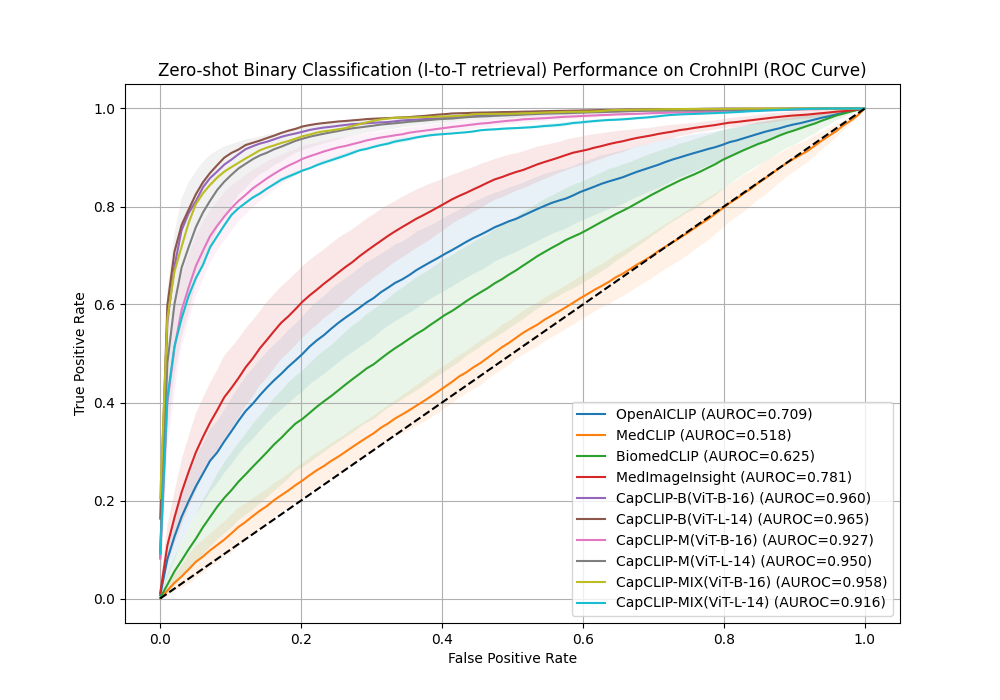}
        \caption{Mean ROC curve across different prompts for zero-shot CLIP-style binary classification on CrohnIPI.}
        \label{fig:crohnipi_roc}
    \end{subfigure}
    \hfill
    \begin{subfigure}[t]{0.7\textwidth}
        \centering
        \includegraphics[width=\linewidth]{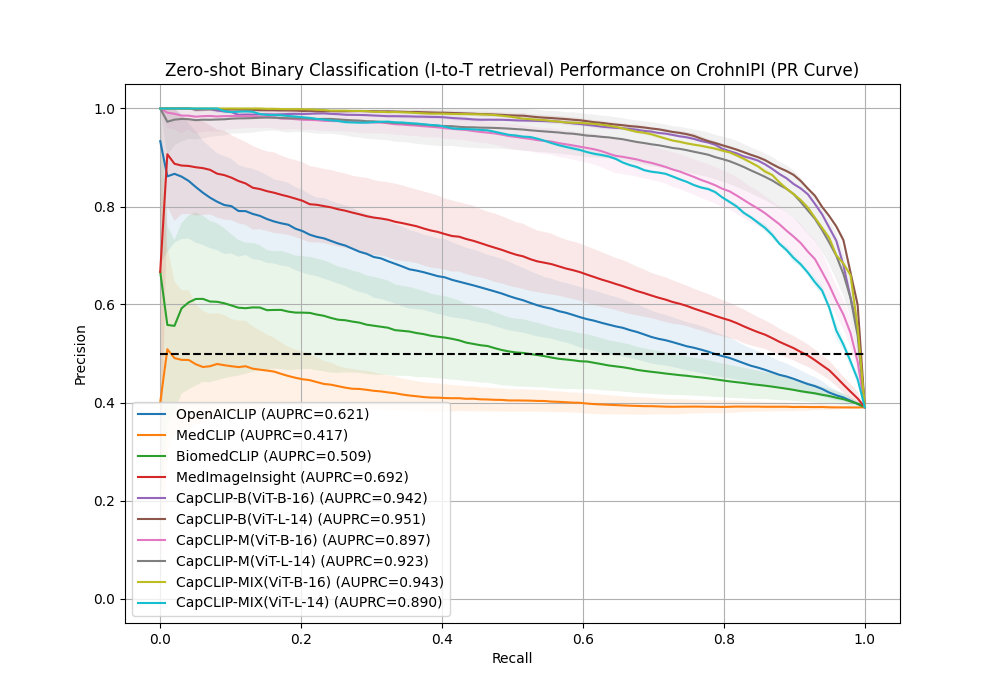}
        \caption{Mean precision-recall curve across different prompts for zero-shot CLIP-style binary classification on CrohnIPI.}
        \label{fig:crohnipi_pr}
    \end{subfigure}
    \caption{Zero-shot CLIP-style binary classification performance on the CrohnIPI dataset. The shaded band indicates variation with respect to prompt sets, and the dotted horizontal line denotes the random baseline.}
    \label{fig:crohnipi_curves}
\end{figure*}

\begin{table*}[t]
\centering
\caption{Zero-shot CLIP-style multiclass image-text classification results. Macro F1 and weighted F1 are reported for each dataset. The best baseline in each column is marked with an asterisk (*), while the best CapCLIP result is shown in bold.}
\label{tab:clip_multiclass}
\renewcommand{\arraystretch}{1.12}
\setlength{\tabcolsep}{5pt}
\footnotesize
\begin{tabular}{lcccccc}
\hline
\textbf{Model} & \multicolumn{2}{c}{\textbf{KID2}} & \multicolumn{2}{c}{\textbf{CrohnIPI}} & \multicolumn{2}{c}{\textbf{Galar}} \\
\cline{2-3} \cline{4-5} \cline{6-7}
 & \textbf{Macro F1} & \textbf{Weighted F1} & \textbf{Macro F1} & \textbf{Weighted F1} & \textbf{Macro F1} & \textbf{Weighted F1} \\
\hline
CLIP-OpenAI         & 0.235 & 0.329 & 0.081 & 0.195* & 0.051 & 0.064* \\
MedCLIP             & 0.204 & 0.287 & 0.095* & 0.185  & 0.047 & 0.057  \\
BioMedCLIP          & 0.316* & 0.437* & 0.075 & 0.078  & 0.040 & 0.035  \\
MedImageInsight     & 0.254 & 0.325 & 0.051 & 0.072  & 0.065* & 0.052  \\
\hline
CapCLIP-B (ViT-B/16)   & 0.328 & 0.460 & 0.151 & 0.526 & 0.067 & 0.161 \\
CapCLIP-B (ViT-L/14)   & 0.331 & 0.488 & 0.178 & 0.577 & 0.084 & 0.172 \\
CapCLIP-M (ViT-B/16)   & 0.442 & 0.551 & 0.208 & 0.495 & 0.110 & 0.198 \\
CapCLIP-M (ViT-L/14)   & \textbf{0.487} & \textbf{0.600} & 0.195 & 0.516 & \textbf{0.143} & \textbf{0.229} \\
CapCLIP-MIX (ViT-B/16) & 0.359 & 0.479 & 0.211 & 0.548 & 0.071 & 0.170 \\
CapCLIP-MIX (ViT-L/14) & 0.389 & 0.448 & \textbf{0.276} & \textbf{0.598} & 0.060 & 0.145 \\
\hline
Improvement over best baseline (\%) & +54.1 & +37.3 & +190.5 & +206.7 & +120.0 & +257.8 \\
\hline
\end{tabular}
\end{table*}

\begin{figure}[t]
    \centering
    \includegraphics[width=\linewidth]{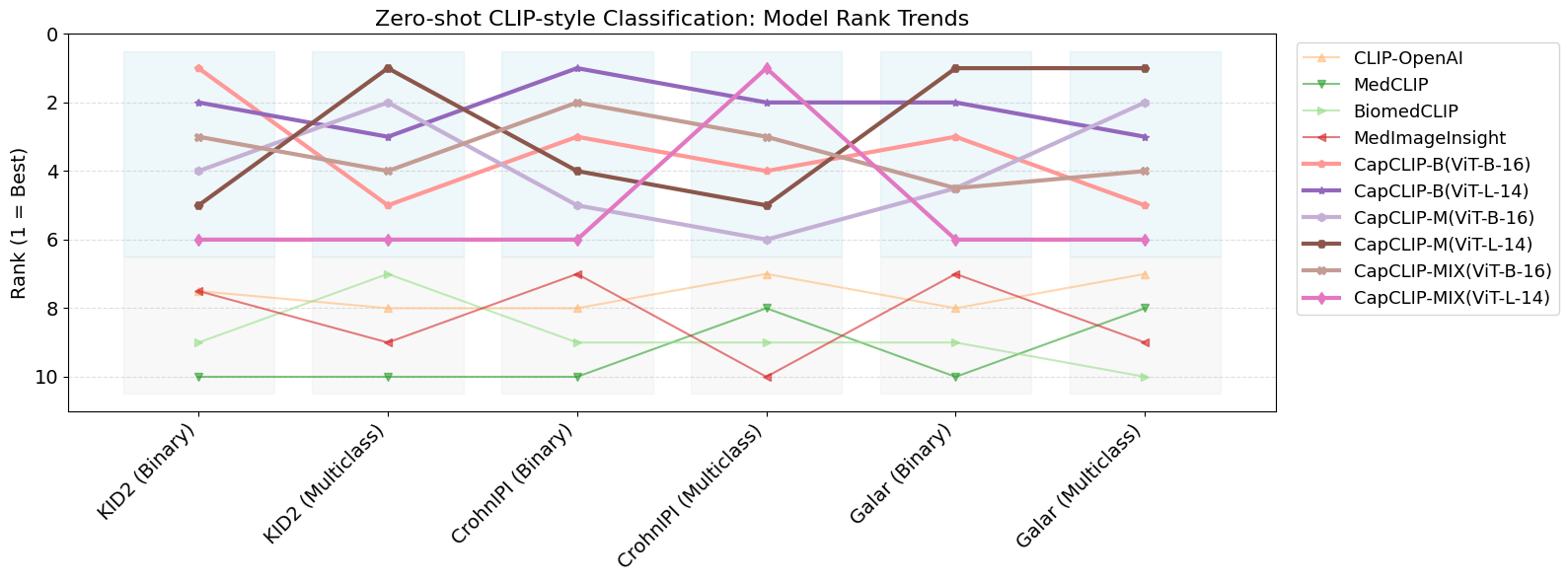}
    \caption{Zero-shot CLIP-style classification/image-text retrieval: model rankings based on weighted F1 across different datasets. CapCLIP models, shown with thicker lines, consistently occupy higher-ranking positions across datasets, whereas the baseline models are concentrated in the lower-ranking region.}
    \label{fig:clipstyle_ranking}
\end{figure}
\subsection{Zero-shot CLIP-style Image-Text Classification}

The results of zero-shot CLIP-style image-text classification are summarised in Tables~\ref{tab:clip_binary} and \ref{tab:clip_multiclass}. Binary performance is reported in terms of mean AUROC and mean AUPRC, averaged across prompt sets, while multiclass performance is reported using macro and weighted F1. The ranking behaviour of all models across datasets and classification modes is shown in Fig.~\ref{fig:clipstyle_ranking}. The CrohnIPI ROC and PR curves are shown in the main text in Fig.~\ref{fig:crohnipi_curves}, while the corresponding ROC and PR curves for KID2 and Galar are provided in Appendix~D.

\textbf{Binary classification.}
CapCLIP outperforms all baseline models by a clear margin across all three datasets in the binary setting. On KID2 and CrohnIPI, AUROC is consistently higher than AUPRC, whereas on Galar the opposite trend is observed. This difference is attributable to class distribution: KID2 and CrohnIPI are relatively more balanced with respect to normal and abnormal classes, while Galar is dominated by abnormal frames. Since AUPRC is more sensitive to performance on the positive class, it becomes comparatively higher when the abnormal class is prevalent.

Among the baselines, MedCLIP and BioMedCLIP perform consistently worse than the general-domain CLIP baseline. This trend mirrors the earlier KNN results and suggests that medical-domain adaptation alone does not guarantee better zero-shot transfer to WCE, especially when the pretraining domain is only weakly aligned with capsule endoscopy imagery. In contrast, MedImageInsight provides the strongest baseline performance across CrohnIPI and Galar, and ties for the best baseline AUPRC on KID2. Nevertheless, all CapCLIP variants surpass the baseline group.

Within the CapCLIP family, the binary model variants are strongest on KID2 and CrohnIPI. In terms of AUPRC, the best CapCLIP result exceeds the best baseline by 54.5\% on KID2 and 37.4\% on CrohnIPI. On Galar, the gap remains substantial, although the strongest result is achieved by CapCLIP-M rather than CapCLIP-B. The CrohnIPI ROC and PR curves shown in Fig.~\ref{fig:crohnipi_curves} further illustrate this separation: CapCLIP shows consistently better curves than the baselines, while also showing lower variation across prompt sets. This relative stability suggests that the text encoder, having been trained on WCE-specific captions and pathology terminology, produces a more aligned semantic space than the off-the-shelf medical or general-domain baselines.

A further observation is that MedCLIP remains close to the random baseline in the curve-based evaluation, indicating weak transfer to WCE despite being a medical-domain model. BioMedCLIP performs better than MedCLIP, but still falls short of CLIP-OpenAI and MedImageInsight. Taken together, these results show that binary image-text classification is a setting in which CapCLIP delivers substantial gains over both general-domain and medical-domain baselines, indicating that domain-specific image-caption alignment provides strong benefits for WCE representation learning.

\textbf{Multiclass classification.}
The multiclass setting is more demanding, as the model must discriminate among multiple pathology categories using only textual prompts. Even under this harder setting, CapCLIP remains consistently stronger than the baselines across all three datasets. On KID2, the best-performing CapCLIP model achieves a macro F1 of 0.487 and a weighted F1 of 0.600, corresponding to improvements of 54.1\% and 37.3\%, respectively, over the strongest baseline. On CrohnIPI, the gains are even larger: the best CapCLIP model reaches a macro F1 of 0.276 and a weighted F1 of 0.598, improving over the strongest baseline by 190.5\% and 206.7\%, respectively.

The Galar dataset remains the most challenging case, owing to its combined semantic and distributional shift. Even so, the best CapCLIP model again achieves the highest macro and weighted F1 scores, with improvements of 120.0\% and 257.8\% over the strongest baseline. Notably, CLIP-OpenAI continues to outperform MedCLIP and BioMedCLIP on several multiclass settings, reinforcing the earlier observation that broad language-supervised pretraining can transfer better to WCE than medical-domain models trained on narrower modalities.

These multiclass results should, however, be interpreted in context. Although the best weighted F1 reaches 0.600, this still reflects a moderately difficult zero-shot problem. Compared with the much stronger binary AUPRC values observed above, the drop in performance is expected, since multiclass classification requires finer semantic discrimination across visually overlapping pathology categories and is more sensitive to prompt phrasing and class imbalance. Nonetheless, the consistent advantage of CapCLIP over both general and medical VLM baselines indicates that WCE-specific multimodal pretraining improves class-level semantic alignment.

The model-ranking trends shown in Fig.~\ref{fig:clipstyle_ranking} provide a complementary view of this behaviour. Across binary and multiclass settings, CapCLIP variants consistently occupy the higher-ranking positions, whereas the baseline models remain concentrated lower in the ranking range. Among the baselines, CLIP-OpenAI shows the most stable behaviour across datasets, followed by MedImageInsight. Within the CapCLIP family, the M variants perform best in most multiclass scenarios, particularly with the ViT-L/14 image encoder, while the B variants are strongest in the binary setting. Overall, these results show that CapCLIP not only improves absolute zero-shot image-text classification performance, but also maintains more favourable ranking behaviour across datasets and evaluation modes.

\begin{table*}[t]
\centering
\caption{Summary of zero-shot abnormality-level coarse text-to-image retrieval performance. The best baseline in each column is marked with an asterisk (*), while the best CapCLIP result is shown in bold. For concise comparison in the main text, mAP and Precision@200 are reported. Detailed per-dataset retrieval results are provided in the appendix.}
\label{tab:retrieval_coarse_summary}
\renewcommand{\arraystretch}{1.12}
\setlength{\tabcolsep}{4.5pt}
\footnotesize
\begin{tabular}{lcccccc}
\hline
\textbf{Model} & \multicolumn{2}{c}{\textbf{KID2}} & \multicolumn{2}{c}{\textbf{CrohnIPI}} & \multicolumn{2}{c}{\textbf{Galar}} \\
\cline{2-3} \cline{4-5} \cline{6-7}
 & \textbf{mAP} & \textbf{P@200} & \textbf{mAP} & \textbf{P@200} & \textbf{mAP} & \textbf{P@200} \\
\hline
CLIP-OpenAI     & 0.438 & 0.440 & 0.448* & 0.493* & 0.738 & 0.715 \\
MedCLIP         & 0.428 & 0.449* & 0.356  & 0.324  & 0.638 & 0.570 \\
BioMedCLIP      & 0.437 & 0.435 & 0.421  & 0.375  & 0.779* & 0.930* \\
MedImageInsight & 0.461* & 0.448 & 0.394  & 0.321  & 0.697 & 0.621 \\
\hline
Best CapCLIP    & \textbf{0.636} & \textbf{0.662} & \textbf{0.870} & \textbf{0.914} & \textbf{0.966} & \textbf{0.999} \\
Variant         & B-L14 & B-L14 & B-L14 & B-L14 & B-L14 & MX-B16 \\
Gain (\%)       & +38.0 & +47.4 & +94.2 & +85.4 & +24.0 & +7.4 \\
\hline
\end{tabular}
\end{table*}

\begin{figure}[t]
    \centering
    \includegraphics[scale=0.4]{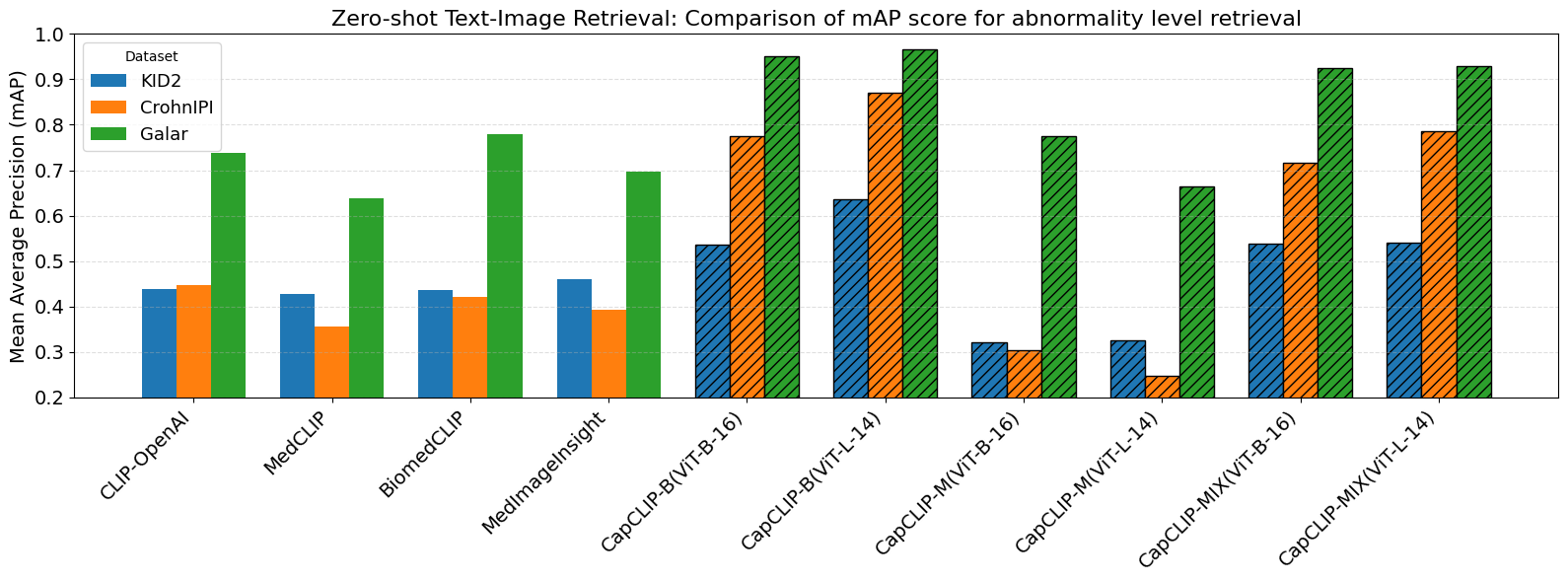}
    \caption{Zero-shot text-to-image retrieval at the abnormality level: comparison of mAP between CapCLIP and the baseline models. CapCLIP variants are shown with striped colours for visual distinction. CapCLIP consistently achieves stronger retrieval performance than the baselines, with the exception of the CapCLIP-M variants.}
    \label{fig:retrieval_coarse_bar}
\end{figure}

\subsection{Zero-shot Text-to-Image Retrieval}

\begin{figure}[t]
    \centering
    \includegraphics[width=\linewidth]{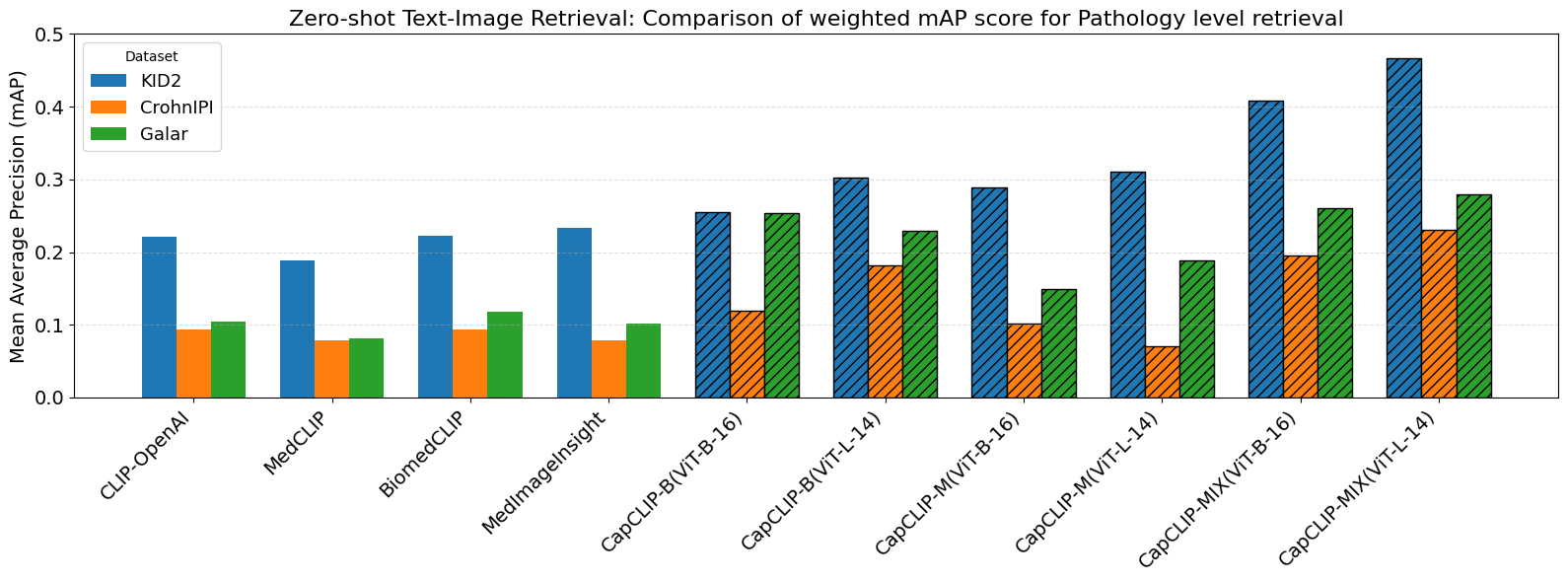}
    \caption{Zero-shot text-to-image retrieval at the pathology level: comparison of weighted mAP between CapCLIP and the baseline models. CapCLIP variants are shown with striped colours for visual distinction.}
    \label{fig:retrieval_fine_bar}
\end{figure}

The results of zero-shot text-to-image retrieval are evaluated under two retrieval modes: abnormality-level coarse retrieval and pathology-level fine-grained retrieval. Summary results for these two settings are presented in Tables~\ref{tab:retrieval_coarse_summary} and \ref{tab:retrieval_fine_summary}, while the corresponding comparisons of mAP and weighted mAP are shown in Figs.~\ref{fig:retrieval_coarse_bar} and \ref{fig:retrieval_fine_bar}, respectively. Model ranking trends across datasets and retrieval modes are illustrated in Fig.~\ref{fig:retrieval_fine_rank}. Detailed per-dataset retrieval results are provided in the appendix.

\textbf{Abnormality-level coarse retrieval.}
In the coarse retrieval setting, textual queries are formulated at the level of abnormality detection rather than specific pathology subtype. All abnormal frames are treated as relevant to the query, whereas normal frames are treated as irrelevant. This corresponds to a binary retrieval problem and reflects the initial screening stage of WCE analysis, where the main objective is to rapidly identify potentially pathological regions within a long video sequence. From a clinical perspective, such a retrieval mode is valuable because it can suppress large volumes of normal frames and bring anomalous frames to the top of the ranked list for further review.

Since abnormality-level queries are not tied to any one dataset or lesion category, a set of 50 textual queries is used across all datasets. These prompts represent different ways in which a clinician might search for abnormal content in WCE. Retrieval metrics are computed for each query and then averaged across the full query set. As summarised in Table~\ref{tab:retrieval_coarse_summary}, CapCLIP shows a clear advantage over the baseline models on all three datasets.

On KID2, the strongest baseline is MedImageInsight, but the best CapCLIP model improves the mAP from 0.461 to 0.636 and Precision@200 from 0.448 to 0.662. On CrohnIPI, CLIP-OpenAI provides the strongest baseline, yet the best CapCLIP result raises mAP from 0.448 to 0.870 and Precision@200 from 0.493 to 0.914. On Galar, BioMedCLIP is the strongest baseline, but CapCLIP again achieves the best overall performance, increasing mAP from 0.779 to 0.966 and Precision@200 from 0.930 to 0.999. These gains indicate that the proposed model retrieves more relevant abnormal frames while also concentrating them more effectively in the highest-ranked positions.

A consistent pattern is also observed across the CapCLIP variants. The binary variants, particularly CapCLIP-B with the ViT-L/14 image encoder, perform best across all datasets, while the mixed variants remain competitive and often provide the second-best results. In contrast, the multiclass-only CapCLIP-M variants perform poorly for this task. This behaviour is coherent with the retrieval objective itself: because the query asks only for abnormal content and does not require subtype discrimination, the binary alignment strategy is better matched to the task than the multiclass formulation.

The detailed results in the appendix further show that the strongest CapCLIP models retrieve substantially larger numbers of abnormal frames within the top-ranked results than the baselines, while maintaining higher precision at the same cut-offs. This is particularly important for WCE, where retrieval usefulness depends not merely on whether abnormal frames are retrieved, but on whether they are brought forward early enough to support efficient first-pass screening. Overall, the coarse retrieval results show that CapCLIP markedly improves the retrieval of abnormal frames compared with both general-domain and medical-domain baselines, with the binary and mixed variants providing the most clinically useful behaviour.

\begin{table*}[t]
\centering
\caption{Summary of zero-shot pathology-level fine-grained text-to-image retrieval performance. The best baseline in each column is marked with an asterisk (*), while the best CapCLIP result is shown in bold. For concise comparison in the main text, macro mAP (M-mAP), weighted mAP (W-mAP), and weighted Precision@200 (W-P@200) are reported. Detailed per-dataset retrieval results are provided in the appendix.}
\label{tab:retrieval_fine_summary}
\renewcommand{\arraystretch}{1.1}
\scriptsize
\resizebox{\textwidth}{!}{%
\begin{tabular}{lccccccccc}
\hline
\textbf{Model} & \multicolumn{3}{c}{\textbf{KID2}} & \multicolumn{3}{c}{\textbf{CrohnIPI}} & \multicolumn{3}{c}{\textbf{Galar}} \\
\cline{2-4} \cline{5-7} \cline{8-10}
 & \textbf{M-mAP} & \textbf{W-mAP} & \textbf{W-P@200}
 & \textbf{M-mAP} & \textbf{W-mAP} & \textbf{W-P@200}
 & \textbf{M-mAP} & \textbf{W-mAP} & \textbf{W-P@200} \\
\hline
CLIP-OpenAI     & 0.177 & 0.221 & 0.225* & 0.080* & 0.093* & 0.101* & 0.087 & 0.104 & 0.073 \\
MedCLIP         & 0.148 & 0.189 & 0.189  & 0.062  & 0.079  & 0.071  & 0.065 & 0.082 & 0.049 \\
BioMedCLIP      & 0.184 & 0.222 & 0.223  & 0.078  & 0.093* & 0.093  & 0.102* & 0.118* & 0.097* \\
MedImageInsight & 0.194* & 0.233* & 0.223  & 0.066  & 0.079  & 0.045  & 0.085 & 0.101 & 0.072 \\
\hline
Best CapCLIP    & \textbf{0.385} & \textbf{0.467} & \textbf{0.479}
                 & \textbf{0.225} & \textbf{0.230} & \textbf{0.216}
                 & \textbf{0.176} & \textbf{0.280} & \textbf{0.350} \\
Variant         & MX-L14 & MX-L14 & MX-L14 & MX-L14 & MX-L14 & MX-L14 & M-L14 & MX-L14 & MX-L14 \\
Gain (\%)       & +98.5 & +100.4 & +112.9 & +181.3 & +147.3 & +113.9 & +72.5 & +137.3 & +260.8 \\
\hline
\end{tabular}%
}
\end{table*}

\textbf{Pathology-level fine-grained retrieval.}
The fine-grained retrieval results are presented here. Here, queries are formulated for specific pathology categories, such as bleeding, polyp, or angiodysplasia, and relevance is defined at the level of the underlying pathology rather than abnormality in general. This corresponds to a multiclass retrieval problem and reflects a later stage of WCE interpretation, where the physician must not only detect abnormality but also distinguish among different lesion types. For each pathology, five different textual queries are used, and the reported scores are averaged across queries.

For this mode, macro and weighted mAP are reported together with weighted Precision@200, as summarised in Table~\ref{tab:retrieval_fine_summary}. Across all three datasets, CapCLIP again outperforms the baseline models, although the absolute scores are lower than those observed for coarse retrieval, reflecting the greater difficulty of fine-grained pathology discrimination.

On KID2, MedImageInsight provides the strongest baseline in terms of both macro and weighted mAP, but the best CapCLIP model, CapCLIP-MIX (ViT-L/14), raises macro mAP from 0.194 to 0.385 and weighted mAP from 0.233 to 0.467. Weighted Precision@200 also improves substantially, from 0.223 for the strongest baseline to 0.479 for the best CapCLIP model. This indicates that the proposed model retrieves a markedly higher concentration of pathology-matched frames within the top ranks.

A similar but even more pronounced pattern is observed on CrohnIPI. Here, the strongest baseline weighted mAP is only 0.093, indicating that the baseline models have very limited ability to separate pathology types in this dataset. The best CapCLIP model raises macro mAP to 0.225 and weighted mAP to 0.230, representing improvements of 181.3\% and 147.3\%, respectively. Weighted Precision@200 increases from 0.101 for the strongest baseline to 0.216 for the best CapCLIP variant, again showing a substantial gain in early-rank retrieval quality.

On Galar, BioMedCLIP forms the strongest baseline, but the best CapCLIP results remain clearly higher. Macro mAP increases from 0.102 to 0.176, weighted mAP from 0.118 to 0.280, and weighted Precision@200 from 0.097 to 0.350. Although the absolute scores remain modest, these improvements show that the proposed domain-specific alignment strategy provides a meaningful advantage in retrieving pathology-specific frames under a challenging evaluation setting.

The overall pattern is shown in Fig.~\ref{fig:retrieval_fine_bar}, where CapCLIP consistently exceeds the baseline group across datasets. In contrast to abnormality-level retrieval, the fine-grained setting is best served by the mixed variants, particularly CapCLIP-MIX with the ViT-L/14 backbone, which gives the strongest overall results in most cases. This suggests that combining binary and multiclass caption supervision is beneficial when the retrieval task requires both broad abnormality awareness and finer pathology-level discrimination.

Finally, the model-ranking trends across both retrieval modes are shown in Fig.~\ref{fig:retrieval_fine_rank}. CapCLIP-B and CapCLIP-MIX occupy the highest-ranking positions overall, with CapCLIP-B performing best for abnormality-level retrieval and CapCLIP-MIX performing best for pathology-level retrieval. In contrast, CapCLIP-M ranks lower in several retrieval settings, indicating that this variant is less suitable when retrieval performance is a primary objective. Taken together, these results show that CapCLIP substantially improves zero-shot text-to-image retrieval for WCE, both as a coarse abnormality filter and as a more demanding fine-grained pathology retrieval framework.

\begin{figure}[t]
    \centering
    \includegraphics[width=\linewidth]{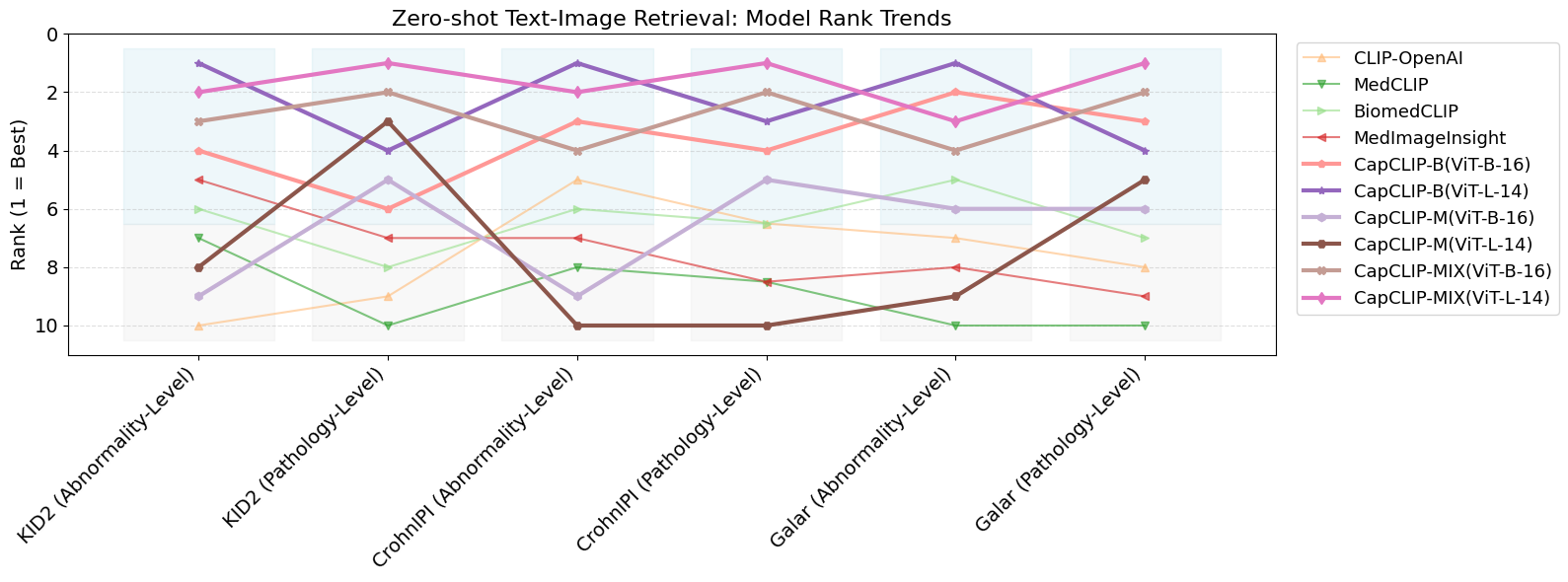}
    \caption{Zero-shot text-to-image retrieval at the pathology level: model rankings based on weighted mAP across different datasets. CapCLIP variants, shown with thicker lines, consistently occupy higher-ranking positions across datasets, whereas the baseline models remain concentrated in the lower-ranking region.}
    \label{fig:retrieval_fine_rank}
\end{figure}

\subsection{Embedding Space Visualisation}

\begin{figure}[t]
    \centering
    \includegraphics[width=0.95\textwidth]{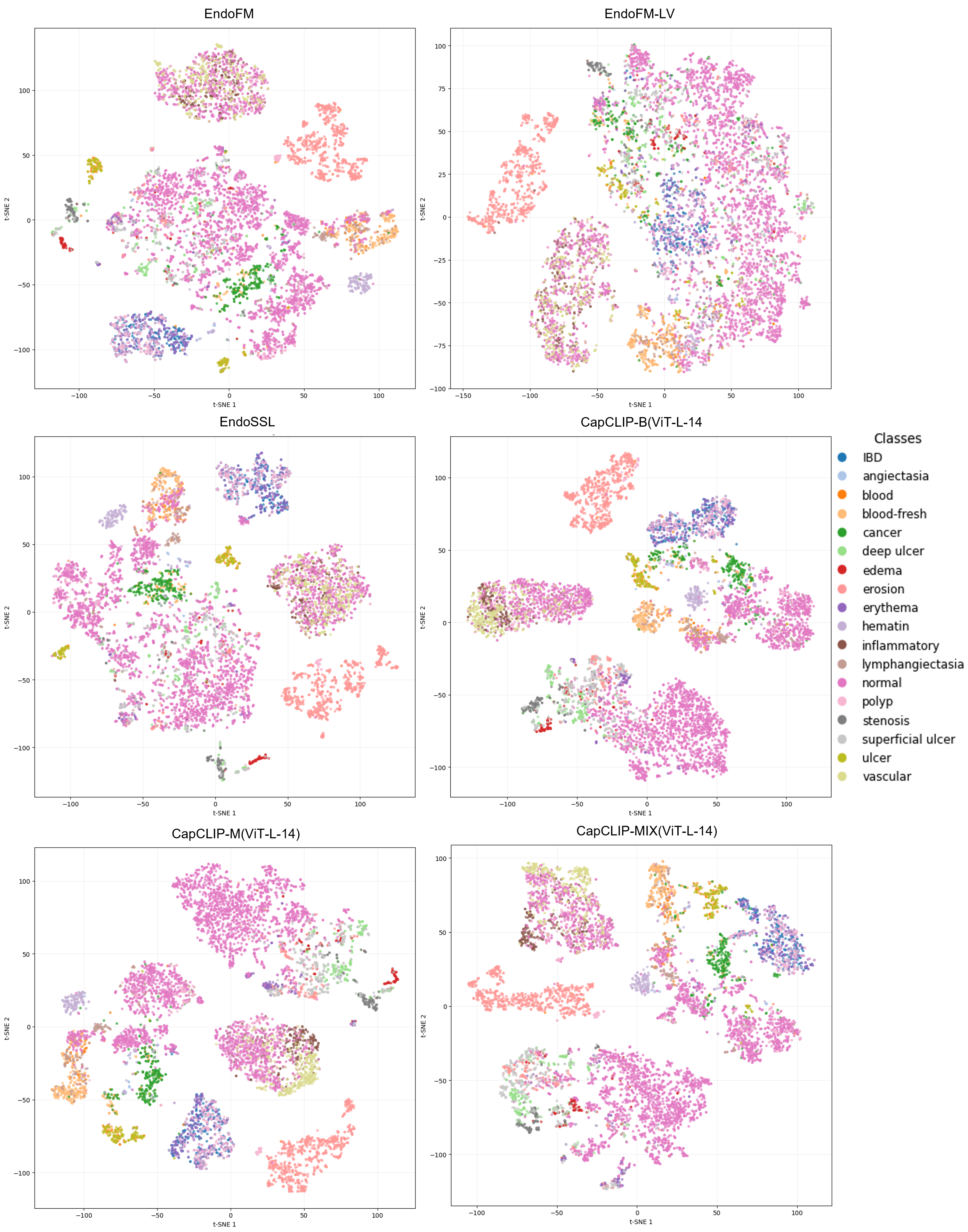}
    \caption{2D t-SNE visualisation of image embeddings from the KID2, CrohnIPI, and Galar datasets. Comparison of CapCLIP with vision-only endoscopic foundation models. CapCLIP variants exhibit clearer clustering structure and stronger inter-class separability than the compared endoscopic baselines.}
    \label{fig:tsne_vision_only}
\end{figure}

\begin{figure}[t]
    \centering
    \includegraphics[width=0.95\textwidth]{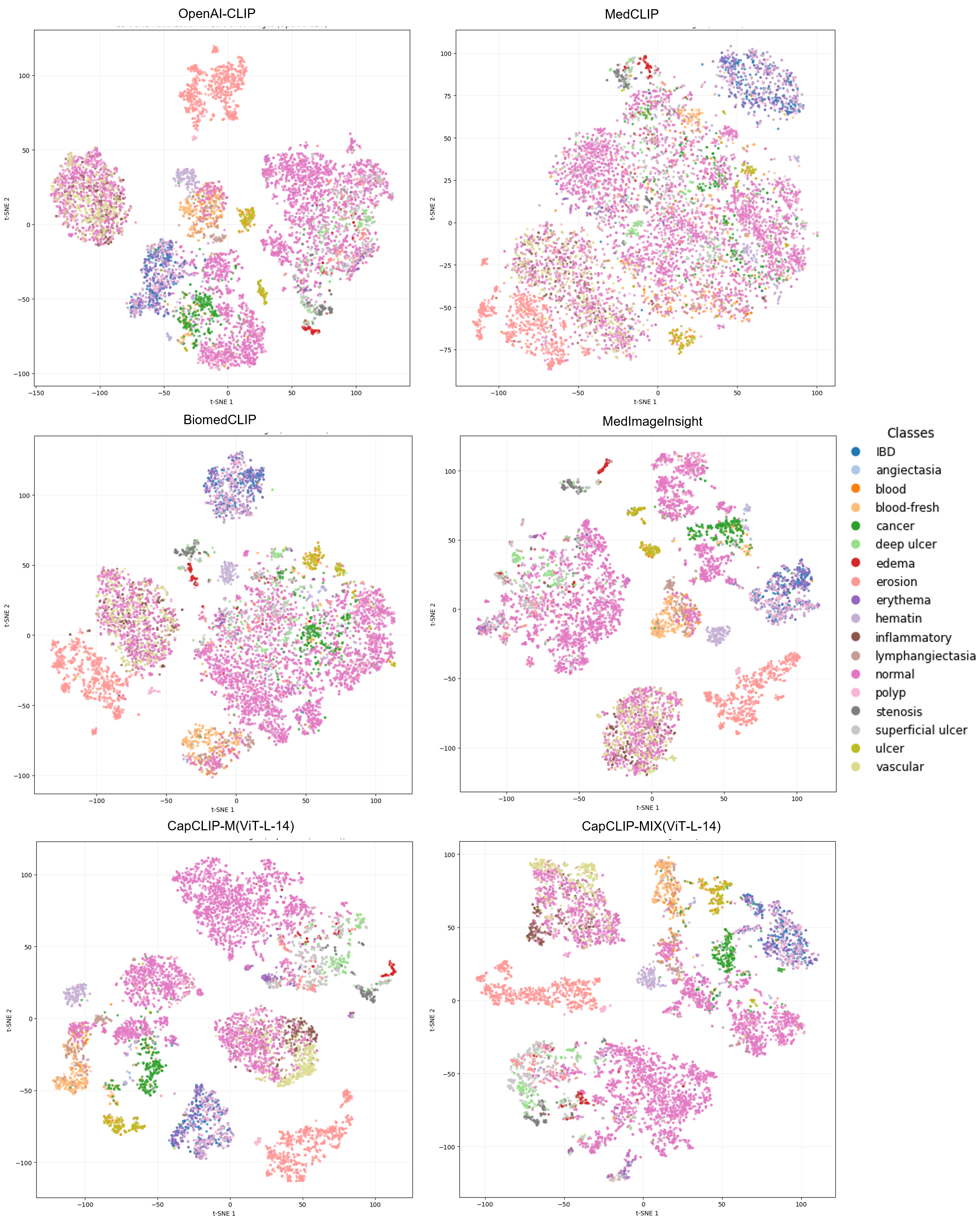}
    \caption{2D t-SNE visualisation of image embeddings from the KID2, CrohnIPI, and Galar datasets. Comparison of CapCLIP with vision-language foundation models.}
    \label{fig:tsne_vlm}
\end{figure}

As an additional qualitative evaluation, the learned image embeddings were projected into two dimensions using t-SNE \cite{tsne}. The resulting visualisations are shown in Figs.~\ref{fig:tsne_vision_only} and \ref{fig:tsne_vlm}, where CapCLIP is compared with (i) vision-only endoscopic foundation models and (ii) medical vision-language foundation models, respectively. Each point corresponds to one image instance, and colours denote the eighteen frame-level finding categories considered in this analysis.

When compared with vision-only endoscopic baselines, CapCLIP produces substantially clearer cluster structure, with improved class-wise grouping and greater separation between categories. In contrast, the endoscopic foundation models show more intermixed distributions, weaker class alignment, and lower visual separability. A similar pattern is observed in comparison with the medical vision-language baselines shown in Fig.~\ref{fig:tsne_vlm}. Among these, MedImageInsight yields the most structured embedding space, while EndoSSL remains the strongest among the vision-only baselines. Nevertheless, the CapCLIP variants exhibit the most coherent and discriminative arrangement overall.

These visualisations provide qualitative support for the quantitative results reported in the previous subsections. In particular, they indicate that CapCLIP learns embeddings that better capture pathology-relevant structure in WCE data. Notably, several categories such as vascular, inflammatory, neoplastic, and IBD-related findings form coherent and separable clusters despite not being explicitly used in the image-caption alignment during training. This suggests that the learned embedding space retains a degree of semantic organisation that extends beyond the directly aligned training categories.

\FloatBarrier

\section{Discussion}
\label{disc}

\subsection{What the Evaluated Tasks Reveal}

The three evaluated tasks provide complementary perspectives on the usefulness of the learned representation space. Although all of them serve as probes of transferability and semantic structure, they differ in how directly they map onto prospective clinical use. Taken together, they show that the benefits of CapCLIP are not limited to one evaluation setting, but extend across visual probing, image-text discrimination, and language-driven retrieval.

The KNN-based image-image retrieval setting provides a useful, though indirect, view of representation quality. While similarity-based image search may have occasional relevance in WCE, it is not the most clinically transformative of the evaluated tasks, since it could in principle also be supported by vision-only systems. Its importance here lies mainly in what it reveals about the embedding space itself: the stronger KNN performance of CapCLIP across most datasets suggests that language-guided supervision improves the semantic organisation of visual features, which in turn benefits other downstream tasks. This interpretation is consistent with the trends shown in Table~\ref{tab:knn_summary}, Fig.~\ref{fig:knngraphics}, and Fig.~\ref{fig:knn_ranktrends}.

CLIP-style image-text classification is more directly relevant to clinical decision support because it parallels the function of existing pathology classifiers, but without being restricted to a fixed closed taxonomy. Conventional classifiers are limited to the categories seen during training and cannot naturally accommodate unseen or rare findings. In contrast, the image-text formulation allows new prompts to be introduced at inference time, making the system more adaptable to semantically varied clinical queries. The strong binary and multiclass results obtained by CapCLIP in Tables~\ref{tab:clip_binary} and \ref{tab:clip_multiclass}, together with the ranking trends in Fig.~\ref{fig:clipstyle_ranking}, suggest that domain-specific image-caption alignment offers a more flexible alternative to rigid label-based classification.

Among the evaluated tasks, text-to-image retrieval is arguably the most aligned with the practical demands of WCE analysis. A typical examination contains tens of thousands of frames, making exhaustive manual review burdensome. In this setting, the ability to issue language-based queries and retrieve relevant frames directly is far more powerful than receiving a static shortlist generated by a fixed classifier. The retrieval results in Tables~\ref{tab:retrieval_coarse_summary} and \ref{tab:retrieval_fine_summary}, together with Figs.~\ref{fig:retrieval_coarse_bar}, \ref{fig:retrieval_fine_bar}, and \ref{fig:retrieval_fine_rank}, show that CapCLIP supports this paradigm effectively, particularly through strong abnormality-level retrieval and clear gains in fine-grained pathology retrieval. This makes text-to-image retrieval the clearest demonstration of how multimodal alignment can support interactive, clinically meaningful exploration of WCE data.

\subsection{Transferability to Unseen Pathologies and Clinical Concepts}

A notable property of CapCLIP in this study is its ability to transfer beyond the pathologies explicitly represented during training. Because the framework aligns images with semantically informative captions rather than relying only on fixed categorical supervision, it can benefit from proxy signals that extend beyond the closed training label space. This allows clinically meaningful concepts to be accessed at inference time through textual prompts, even when their visual instances were limited or absent during training.

The dataset distributions used in this work illustrate the importance of this property. As shown by the training and test class distributions in Appendix~B, several findings are either underrepresented or not aligned directly in the training setup, yet coherent performance is still observed on the unseen evaluation datasets. This behaviour is also qualitatively supported by the embedding visualisations in Figs.~\ref{fig:tsne_vision_only} and \ref{fig:tsne_vlm}, where categories such as vascular, inflammatory, neoplastic, and IBD-related findings form more coherent structures under CapCLIP than under the compared baselines. In a conventional vision-only classifier, such categories would typically be forced into the closest known class, limiting both flexibility and clinical usefulness.

More broadly, this behaviour reflects a key advantage of vision-language learning: transfer is mediated not only by visual similarity, but also by semantic relationships carried through language. While this is a recognised property of multimodal foundation models in the broader literature \cite{radford2021learning, med-flamingo, zhang2023biomedclip}, its demonstration in WCE is particularly important. Capsule endoscopy often involves rare, subtle, and clinically diverse findings, and these are precisely the cases in which rigid label-constrained classifiers are least adequate. In this setting, the ability to extend to new pathology concepts without retraining or manual relabelling represents a meaningful step toward more adaptable clinical AI systems.

\subsection{Rethinking Conventional Approaches to WCE Analysis}

Much of the existing work on automated WCE analysis remains narrow in scope relative to the requirements of a clinically useful prospective system. Many methods are designed for a single task, such as binary abnormality detection or segmentation of a specific structure, while others are limited to a small subset of pathology categories. Although such approaches have demonstrated the feasibility of applying machine learning to WCE, their practical flexibility remains limited. Even broader taxonomy-based classification frameworks \cite{Wahab_hollistic}, while extending anomaly coverage, still operate within fixed label spaces and do not naturally support transfer across tasks or semantically richer forms of interaction.

This limitation is also reflected in current clinical support tools, which typically rely on deep learning models to prioritise frames likely to contain abnormalities or to assign predictions within a predefined taxonomy. While such systems may reduce reading time, they remain largely confined to closed-set decision support and provide limited transparency regarding how or why a given result has been produced. As a result, they are more often used as supplementary aids than as fully trusted diagnostic tools \cite{wahab2023}.

A next-generation WCE analysis system should therefore move beyond rigid task and category-specific prediction toward flexible, semantically grounded interaction. In practice, this means enabling physicians to interrogate long video sequences not only through predefined categories, but also through clinically meaningful language-based concepts. Since clinical reasoning is naturally expressed in language, grounding visual representations in text offers a more intuitive route toward interpretability, retrieval, and user trust. From this perspective, cross-modal alignment is not only a technical design choice, but also a practical step toward building WCE analysis systems that better reflect the way clinicians reason, explore, and validate visual evidence.

\section{Conclusion}

This paper investigated vision-language representation learning for wireless capsule endoscopy and evaluated both existing multimodal foundation models and the proposed CapCLIP framework under out-of-distribution conditions. Across KNN probing, CLIP-style image-text classification, and text-to-image retrieval, CapCLIP consistently outperformed the compared baselines, demonstrating that domain-specific image-caption alignment yields more transferable and semantically useful representations for WCE than both general-domain and medical-domain alternatives. Among the baseline models, MedImageInsight provided the strongest overall medical-domain performance, while CLIP-OpenAI remained a competitive general-domain baseline. Overall, the results support the view that WCE-specific vision-language pretraining offers a more effective path toward flexible and clinically meaningful analysis than conventional fixed-label approaches.

Several limitations remain. CapCLIP relies on generated captions rather than naturally paired clinical text, and the scale of available WCE training data remains far smaller than that used for a typical foundation model. In addition, different CapCLIP variants were found to be better suited to different downstream tasks, indicating that a single universally optimal configuration was not achieved in the present study. Future work should therefore focus on larger and more diverse WCE data, stronger unified training strategies that combine the strengths of the current variants, and evaluation under broader clinical settings.

\FloatBarrier

\appendix
\counterwithin{figure}{section}
\counterwithin{table}{section}
\renewcommand{\thefigure}{\thesection\arabic{figure}}
\renewcommand{\thetable}{\thesection\arabic{table}}
\section{Caption Generation Details}
\label{app:caption_details}
\begin{figure}[t]
    \centering
    \includegraphics[scale=0.75]{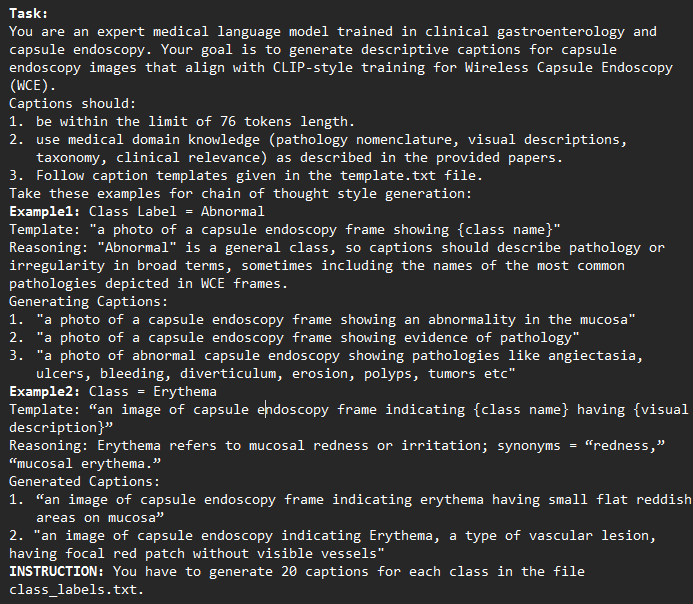}
    \caption{Prompt used for caption generation.}
    \label{fig:appendix_prompt}
\end{figure}

The prompt used for caption generation is shown in Figure \ref{fig:appendix_prompt}. Some examples of captions used are presented in Table \ref{tab:appendix_captions}.

\begin{table*}[t]
\centering
\caption{Some examples of generated captions for the classes of Erythema, Abnormal, Erosion, and Normal.}
\label{tab:appendix_captions}
\renewcommand{\arraystretch}{1.1}
\setlength{\tabcolsep}{9pt}
\tiny
\begin{tabular}{p{2.5cm} p{13cm}}
\hline
\textbf{Class Name} & \textbf{Generated Captions} \\
\hline

\textbf{Erythema} &
``a photo of an abnormal capsule endoscopy frame showing Erythema (erythematous patch), a vascular lesion, small flat reddish area on mucosa'' \\

& ``a photo of an abnormal SBCE frame: Erythema (erythematous patch), vascular lesion, small flat red area on a fold'' \\

& ``an abnormal image of capsule endoscopy depicting Erythema, under the vascular lesion parent'' \\
\hline

\textbf{Abnormal} &
``a photo of an abnormal capsule endoscopy image showing pathology or anomaly.'' \\

& ``a photo of an abnormal capsule endoscopy image showing any one of the following pathologies such as Aphthous erosion, superficial or deep ulceration, Edema, erythema, stenosis with possible clinical relevance to Crohn's disease.'' \\

& ``a photo of an abnormal capsule endoscopy image showing pathology or anomaly such as polyps, tumours, blood, angiectasia, lymphatic lesions, or vascular lesions'' \\
\hline

\textbf{Erosion} &
``an abnormal capsule endoscopy frame labelled Erosion or Aphthae, a type of inflammatory/ulcerative lesion'' \\

& ``an abnormal capsule endoscopy frame of Erosion or Aphthae, inflammatory/ulcerative lesion, several millimetric erosions arranged linearly along a fold'' \\

& ``a pathology image from capsule endoscopy with Erosion or Aphthae, inflammatory/ulcerative lesion, small whitish aphthoid lesion bordered by erythema'' \\
\hline

\textbf{Normal} &
``an image of capsule endoscopy with normal small bowel appearance, pink villous surface without lesions'' \\

& ``a photo of a normal capsule endoscopy frame showing healthy small bowel mucosa with regular circular folds'' \\

& ``a photo of a normal capsule endoscopy image, featuring a pinkish or pinkish-tan surface with a glossy sheen from moisture, and fine folds without lesions or discolouration" \\
\hline

\end{tabular}
\end{table*}

\section{Dataset and Class Distribution Details}
\label{sec:appendix_class_dist}
\begin{figure}[t]
    \centering
    \begin{subfigure}[t]{0.695\textwidth}
        \centering
        \includegraphics[width=\linewidth]{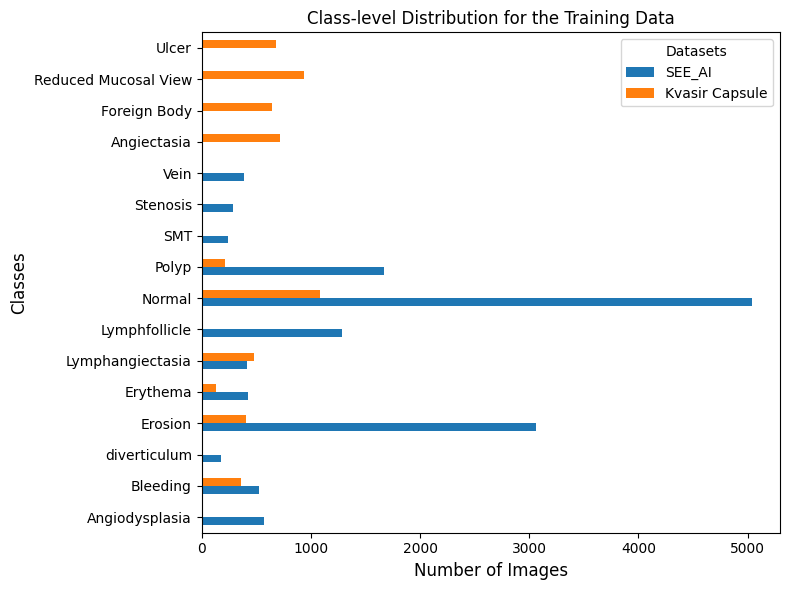}
        \caption{Training dataset class distribution.}
        \label{fig:appendix_training_dist}
    \end{subfigure}
    \hfill
    \begin{subfigure}[t]{0.695\textwidth}
        \centering
        \includegraphics[width=\linewidth]{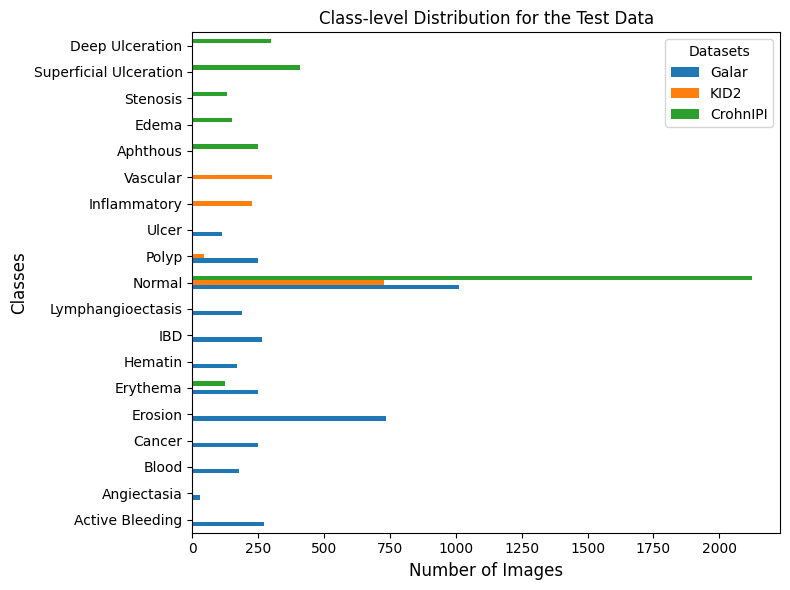}
        \caption{Testing dataset class distribution.}
        \label{fig:appendix_testing_dist}
    \end{subfigure}
    \caption{Pathology class distributions used in the experiments.}
    \label{fig:appendix_distributions}
\end{figure}

Here, Pathology distribution across training and testing datasets is presented in Figure \ref{fig:appendix_training_dist} and \ref{fig:appendix_testing_dist}.

\section{Detailed KNN Classification Results}
\label{sec:appendix_knn}

This appendix provides the detailed zero-shot KNN classification results corresponding to the summary results presented in the main text. Binary classification reports for KID2, CrohnIPI, and Galar are given in Tables~\ref{tab:appendix_knn_kid2_binary}--\ref{tab:appendix_knn_galar_binary}, while multiclass classification results for the same datasets are reported in Tables~\ref{tab:appendix_knn_kid2_multi}--\ref{tab:appendix_knn_galar_multi}. These tables include the full per-class and averaged metrics used to support the comparative discussion in Section~V.

\begin{table*}[tbp]
\centering
\caption{Zero-shot KNN binary classification results for the KID2 dataset. The best baseline macro and weighted F1 scores are marked with an asterisk (*). Percentage improvement over the best baseline is shown in parentheses for CapCLIP models.}
\label{tab:appendix_knn_kid2_binary}
\renewcommand{\arraystretch}{1.1}
\setlength{\tabcolsep}{12pt}
\tiny
\begin{tabular}{lcccccccc}
\hline
\textbf{Model} &
\multicolumn{3}{c}{\textbf{Class Abnormal}} &
\multicolumn{3}{c}{\textbf{Class Normal}} &
\textbf{Macro Avg.} &
\textbf{Weighted Avg.} \\
\cline{2-4} \cline{5-7}
& \textbf{P} & \textbf{R} & \textbf{F1}
& \textbf{P} & \textbf{R} & \textbf{F1}
& \textbf{F1} & \textbf{F1} \\
\hline
EndoFM                & 0.753 & 0.787 & 0.770 & 0.826 & 0.797 & 0.811 & 0.791  & 0.793  \\
EndoFM-LV             & 0.769 & 0.638 & 0.697 & 0.748 & 0.849 & 0.795 & 0.746  & 0.752  \\
EndoSSL               & 0.743 & 0.770 & 0.756 & 0.813 & 0.790 & 0.801 & 0.779  & 0.781  \\
CLIP-OpenAI           & 0.798 & 0.751 & 0.774 & 0.812 & 0.850 & 0.831 & 0.802* & 0.806* \\
MedCLIP               & 0.721 & 0.645 & 0.681 & 0.741 & 0.804 & 0.771 & 0.726  & 0.731  \\
BioMedCLIP            & 0.826 & 0.695 & 0.755 & 0.786 & 0.885 & 0.833 & 0.794  & 0.798  \\
MedImageInsight       & 0.826 & 0.655 & 0.731 & 0.766 & 0.891 & 0.824 & 0.777  & 0.783  \\
CapCLIP-B (ViT-B/16)  & 0.888 & 0.829 & 0.858 & 0.872 & 0.918 & 0.894 & 0.876 (+9.2\%) & 0.878 (+8.9\%) \\
CapCLIP-B (ViT-L/14)  & 0.871 & 0.847 & 0.859 & 0.882 & 0.901 & 0.891 & 0.875 (+9.1\%) & 0.877 (+8.8\%) \\
CapCLIP-M (ViT-B/16)  & 0.903 & 0.796 & 0.846 & 0.853 & 0.933 & 0.891 & 0.869 (+8.4\%) & 0.871 (+8.1\%) \\
CapCLIP-M (ViT-L/14)  & 0.898 & 0.801 & 0.847 & 0.856 & 0.929 & 0.891 & 0.869 (+8.4\%) & 0.872 (+8.2\%) \\
CapCLIP-MIX (ViT-B/16)& 0.877 & 0.817 & 0.846 & 0.863 & 0.909 & 0.886 & 0.866 (+8.0\%) & 0.868 (+7.7\%) \\
CapCLIP-MIX (ViT-L/14)& 0.880 & 0.768 & 0.820 & 0.834 & 0.918 & 0.874 & 0.847 (+5.6\%) & 0.850 (+5.5\%) \\
\hline
\end{tabular}%
\end{table*}

\begin{table*}[tbp]
\centering
\caption{Zero-shot KNN binary classification results for the CrohnIPI dataset. The best baseline macro and weighted F1 scores are marked with an asterisk (*). Percentage improvement over the best baseline is shown in parentheses for CapCLIP models.}
\label{tab:appendix_knn_crohnipi_binary}
\renewcommand{\arraystretch}{1.1}
\setlength{\tabcolsep}{12pt}
\tiny
\begin{tabular}{lcccccccc}
\hline
\textbf{Model} &
\multicolumn{3}{c}{\textbf{Class Abnormal}} &
\multicolumn{3}{c}{\textbf{Class Normal}} &
\textbf{Macro Avg.} &
\textbf{Weighted Avg.} \\
\cline{2-4} \cline{5-7}
& \textbf{P} & \textbf{R} & \textbf{F1}
& \textbf{P} & \textbf{R} & \textbf{F1}
& \textbf{F1} & \textbf{F1} \\
\hline
EndoFM                & 0.901 & 0.919 & 0.910 & 0.948 & 0.935 & 0.941 & 0.926* & 0.929* \\
EndoFM-LV             & 0.882 & 0.856 & 0.869 & 0.909 & 0.927 & 0.918 & 0.893  & 0.899  \\
EndoSSL               & 0.903 & 0.913 & 0.908 & 0.944 & 0.937 & 0.940 & 0.924  & 0.928  \\
CLIP-OpenAI           & 0.864 & 0.893 & 0.878 & 0.930 & 0.910 & 0.920 & 0.899  & 0.904  \\
MedCLIP               & 0.846 & 0.815 & 0.830 & 0.884 & 0.905 & 0.895 & 0.862  & 0.870  \\
BioMedCLIP            & 0.889 & 0.898 & 0.894 & 0.934 & 0.928 & 0.931 & 0.912  & 0.917  \\
MedImageInsight       & 0.891 & 0.925 & 0.908 & 0.951 & 0.927 & 0.939 & 0.923  & 0.927  \\
CapCLIP-B (ViT-B/16)  & 0.916 & 0.943 & 0.929 & 0.963 & 0.945 & 0.954 & 0.941 (+1.6\%) & 0.944 (+1.6\%) \\
CapCLIP-B (ViT-L/14)  & 0.921 & 0.949 & 0.935 & 0.967 & 0.948 & 0.957 & 0.946 (+2.2\%) & 0.949 (+2.2\%) \\
CapCLIP-M (ViT-B/16)  & 0.939 & 0.955 & 0.947 & 0.971 & 0.960 & 0.966 & 0.956 (+3.2\%) & 0.958 (+3.1\%) \\
CapCLIP-M (ViT-L/14)  & 0.955 & 0.965 & 0.960 & 0.977 & 0.971 & 0.974 & 0.967 (+4.4\%) & 0.968 (+4.2\%) \\
CapCLIP-MIX (ViT-B/16)& 0.945 & 0.950 & 0.948 & 0.968 & 0.965 & 0.966 & 0.957 (+3.3\%) & 0.959 (+3.2\%) \\
CapCLIP-MIX (ViT-L/14)& 0.937 & 0.939 & 0.938 & 0.961 & 0.960 & 0.960 & 0.949 (+2.5\%) & 0.952 (+2.5\%) \\
\hline
\end{tabular}%
\end{table*}

\begin{table*}[tbp]
\centering
\caption{Zero-shot KNN binary classification results for the Galar dataset. The best baseline macro and weighted F1 scores are marked with an asterisk (*). Percentage improvement over the best baseline is shown in parentheses for CapCLIP models.}
\label{tab:appendix_knn_galar_binary}
\renewcommand{\arraystretch}{1.1}
\setlength{\tabcolsep}{12pt}
\tiny
\begin{tabular}{lcccccccc}
\hline
\textbf{Model} &
\multicolumn{3}{c}{\textbf{Class Abnormal}} &
\multicolumn{3}{c}{\textbf{Class Normal}} &
\textbf{Macro Avg.} &
\textbf{Weighted Avg.} \\
\cline{2-4} \cline{5-7}
& \textbf{P} & \textbf{R} & \textbf{F1}
& \textbf{P} & \textbf{R} & \textbf{F1}
& \textbf{F1} & \textbf{F1} \\
\hline
EndoFM                & 0.986 & 0.955 & 0.970 & 0.889 & 0.964 & 0.925 & 0.948  & 0.958  \\
EndoFM-LV             & 0.973 & 0.958 & 0.965 & 0.893 & 0.930 & 0.911 & 0.938  & 0.950  \\
EndoSSL               & 0.986 & 0.958 & 0.972 & 0.897 & 0.963 & 0.929 & 0.950  & 0.960  \\
CLIP-OpenAI           & 0.982 & 0.959 & 0.970 & 0.897 & 0.954 & 0.925 & 0.948  & 0.958  \\
MedCLIP               & 0.964 & 0.919 & 0.941 & 0.810 & 0.909 & 0.856 & 0.899  & 0.918  \\
BioMedCLIP            & 0.983 & 0.947 & 0.964 & 0.871 & 0.955 & 0.911 & 0.938  & 0.950  \\
MedImageInsight       & 0.987 & 0.963 & 0.975 & 0.907 & 0.965 & 0.935 & 0.955* & 0.964* \\
CapCLIP-B (ViT-B/16)  & 0.984 & 0.963 & 0.973 & 0.906 & 0.957 & 0.931 & 0.952 (-0.3\%) & 0.962 (-0.2\%) \\
CapCLIP-B (ViT-L/14)  & 0.981 & 0.966 & 0.974 & 0.913 & 0.951 & 0.932 & 0.953 (-0.2\%) & 0.962 (-0.2\%) \\
CapCLIP-M (ViT-B/16)  & 0.991 & 0.962 & 0.976 & 0.907 & 0.976 & 0.940 & 0.958 (+0.3\%) & 0.967 (+0.3\%) \\
CapCLIP-M (ViT-L/14)  & 0.988 & 0.966 & 0.977 & 0.914 & 0.969 & 0.941 & 0.959 (+0.4\%) & 0.967 (+0.3\%) \\
CapCLIP-MIX (ViT-B/16)& 0.982 & 0.949 & 0.965 & 0.875 & 0.954 & 0.913 & 0.939 (-1.7\%) & 0.951 (-1.3\%) \\
CapCLIP-MIX (ViT-L/14)& 0.984 & 0.960 & 0.972 & 0.900 & 0.957 & 0.928 & 0.950 (-0.5\%) & 0.960 (-0.4\%) \\
\hline
\end{tabular}%
\end{table*}

\begin{table*}[tbp]
\centering
\caption{Zero-shot KNN multiclass classification results for the KID2 dataset. The best baseline macro and weighted F1 scores are marked with an asterisk (*). Percentage improvement over the best baseline is shown in parentheses for CapCLIP models.}
\label{tab:appendix_knn_kid2_multi}
\renewcommand{\arraystretch}{1.1}
\setlength{\tabcolsep}{12pt}
\tiny
\begin{tabular}{lcccccccc}
\hline
\textbf{Model} &
\multicolumn{3}{c}{\textbf{Macro Average}} &
\multicolumn{3}{c}{\textbf{Weighted Average}} \\
\cline{2-4} \cline{5-7}
& \textbf{P} & \textbf{R} & \textbf{F1}
& \textbf{P} & \textbf{R} & \textbf{F1} \\
\hline
EndoFM                & 0.744 & 0.621 & 0.651  & 0.759 & 0.751 & 0.745  \\
EndoFM-LV             & 0.639 & 0.476 & 0.499  & 0.689 & 0.699 & 0.674  \\
EndoSSL               & 0.804 & 0.627 & 0.672* & 0.763 & 0.760 & 0.753* \\
CLIP-OpenAI           & 0.744 & 0.561 & 0.588  & 0.761 & 0.750 & 0.737  \\
MedCLIP               & 0.597 & 0.454 & 0.474  & 0.664 & 0.666 & 0.639  \\
BioMedCLIP            & 0.782 & 0.574 & 0.623  & 0.750 & 0.729 & 0.713  \\
MedImageInsight       & 0.803 & 0.564 & 0.617  & 0.761 & 0.733 & 0.714  \\
CapCLIP-B (ViT-B/16)  & 0.838 & 0.680 & 0.718 (+6.8\%) & 0.845 & 0.840 & 0.834 (+10.8\%) \\
CapCLIP-B (ViT-L/14)  & 0.861 & 0.666 & 0.697 (+3.7\%) & 0.843 & 0.829 & 0.822 (+9.2\%) \\
CapCLIP-M (ViT-B/16)  & 0.855 & 0.653 & 0.687 (+2.2\%) & 0.850 & 0.848 & 0.837 (+11.2\%) \\
CapCLIP-M (ViT-L/14)  & 0.823 & 0.681 & 0.722 (+7.4\%) & 0.843 & 0.844 & 0.837 (+11.2\%) \\
CapCLIP-MIX (ViT-B/16)& 0.844 & 0.665 & 0.698 (+3.9\%) & 0.838 & 0.834 & 0.827 (+9.8\%) \\
CapCLIP-MIX (ViT-L/14)& 0.820 & 0.654 & 0.696 (+3.6\%) & 0.829 & 0.825 & 0.816 (+8.4\%) \\
\hline
\end{tabular}%
\end{table*}

\begin{table*}[tbp]
\centering
\caption{Zero-shot KNN multiclass classification results for the CrohnIPI dataset. The best baseline macro and weighted F1 scores are marked with an asterisk (*). Percentage improvement over the best baseline is shown in parentheses for CapCLIP models.}
\label{tab:appendix_knn_crohnipi_multi}
\renewcommand{\arraystretch}{1.1}
\setlength{\tabcolsep}{12pt}
\tiny
\begin{tabular}{lcccccccc}
\hline
\textbf{Model} &
\multicolumn{3}{c}{\textbf{Macro Average}} &
\multicolumn{3}{c}{\textbf{Weighted Average}} \\
\cline{2-4} \cline{5-7}
& \textbf{P} & \textbf{R} & \textbf{F1}
& \textbf{P} & \textbf{R} & \textbf{F1} \\
\hline
EndoFM                & 0.801 & 0.791 & 0.794  & 0.873 & 0.874 & 0.872  \\
EndoFM-LV             & 0.765 & 0.702 & 0.727  & 0.830 & 0.834 & 0.826  \\
EndoSSL               & 0.818 & 0.789 & 0.802  & 0.877 & 0.877 & 0.876* \\
CLIP-OpenAI           & 0.763 & 0.724 & 0.740  & 0.838 & 0.843 & 0.838  \\
MedCLIP               & 0.743 & 0.653 & 0.686  & 0.800 & 0.802 & 0.792  \\
BioMedCLIP            & 0.787 & 0.739 & 0.759  & 0.850 & 0.853 & 0.849  \\
MedImageInsight       & 0.805 & 0.804 & 0.803* & 0.877 & 0.876 & 0.875  \\
CapCLIP-B (ViT-B/16)  & 0.805 & 0.793 & 0.798 (-0.5\%) & 0.881 & 0.881 & 0.880 (+0.5\%) \\
CapCLIP-B (ViT-L/14)  & 0.809 & 0.798 & 0.803 (0.0\%)  & 0.889 & 0.890 & 0.890 (+1.6\%) \\
CapCLIP-M (ViT-B/16)  & 0.828 & 0.819 & 0.823 (+2.5\%) & 0.900 & 0.900 & 0.900 (+2.7\%) \\
CapCLIP-M (ViT-L/14)  & 0.862 & 0.852 & 0.856 (+6.6\%) & 0.919 & 0.920 & 0.919 (+4.9\%) \\
CapCLIP-MIX (ViT-B/16)& 0.831 & 0.827 & 0.828 (+3.1\%) & 0.902 & 0.903 & 0.902 (+3.0\%) \\
CapCLIP-MIX (ViT-L/14)& 0.793 & 0.785 & 0.788 (-1.9\%) & 0.881 & 0.883 & 0.881 (+0.6\%) \\
\hline
\end{tabular}%
\end{table*}

\begin{table*}[tbp]
\centering
\caption{Zero-shot KNN multiclass classification results for the Galar dataset. The best baseline macro and weighted F1 scores are marked with an asterisk (*). Percentage improvement over the best baseline is shown in parentheses for CapCLIP models.}
\label{tab:appendix_knn_galar_multi}
\renewcommand{\arraystretch}{1.1}
\setlength{\tabcolsep}{12pt}
\tiny
\begin{tabular}{lcccccccc}
\hline
\textbf{Model} &
\multicolumn{3}{c}{\textbf{Macro Average}} &
\multicolumn{3}{c}{\textbf{Weighted Average}} \\
\cline{2-4} \cline{5-7}
& \textbf{P} & \textbf{R} & \textbf{F1}
& \textbf{P} & \textbf{R} & \textbf{F1} \\
\hline
EndoFM                & 0.756 & 0.726 & 0.732* & 0.813 & 0.816 & 0.808* \\
EndoFM-LV             & 0.710 & 0.674 & 0.680  & 0.781 & 0.783 & 0.773  \\
EndoSSL               & 0.751 & 0.719 & 0.724  & 0.813 & 0.813 & 0.804  \\
CLIP-OpenAI           & 0.742 & 0.701 & 0.710  & 0.797 & 0.799 & 0.790  \\
MedCLIP               & 0.697 & 0.629 & 0.647  & 0.748 & 0.749 & 0.735  \\
BioMedCLIP            & 0.733 & 0.674 & 0.689  & 0.794 & 0.795 & 0.785  \\
MedImageInsight       & 0.765 & 0.715 & 0.725  & 0.818 & 0.815 & 0.806  \\
CapCLIP-B (ViT-B/16)  & 0.747 & 0.708 & 0.718 (-1.9\%) & 0.808 & 0.809 & 0.802 (-0.7\%) \\
CapCLIP-B (ViT-L/14)  & 0.746 & 0.724 & 0.726 (-0.8\%) & 0.810 & 0.812 & 0.804 (-0.5\%) \\
CapCLIP-M (ViT-B/16)  & 0.743 & 0.714 & 0.718 (-1.9\%) & 0.809 & 0.812 & 0.803 (-0.6\%) \\
CapCLIP-M (ViT-L/14)  & 0.748 & 0.724 & 0.729 (-0.4\%) & 0.812 & 0.813 & 0.806 (-0.2\%) \\
CapCLIP-MIX (ViT-B/16)& 0.719 & 0.679 & 0.687 (-6.1\%) & 0.794 & 0.795 & 0.785 (-2.8\%) \\
CapCLIP-MIX (ViT-L/14)& 0.739 & 0.713 & 0.715 (-2.3\%) & 0.807 & 0.804 & 0.797 (-1.4\%) \\
\hline
\end{tabular}%
\end{table*}

\section{Detailed CLIP-style Classification Results}
\label{sec:appendix_clipstyle}
\begin{figure*}[t]
    \centering
    \begin{subfigure}[t]{0.7\textwidth}
        \centering
        \includegraphics[width=\linewidth]{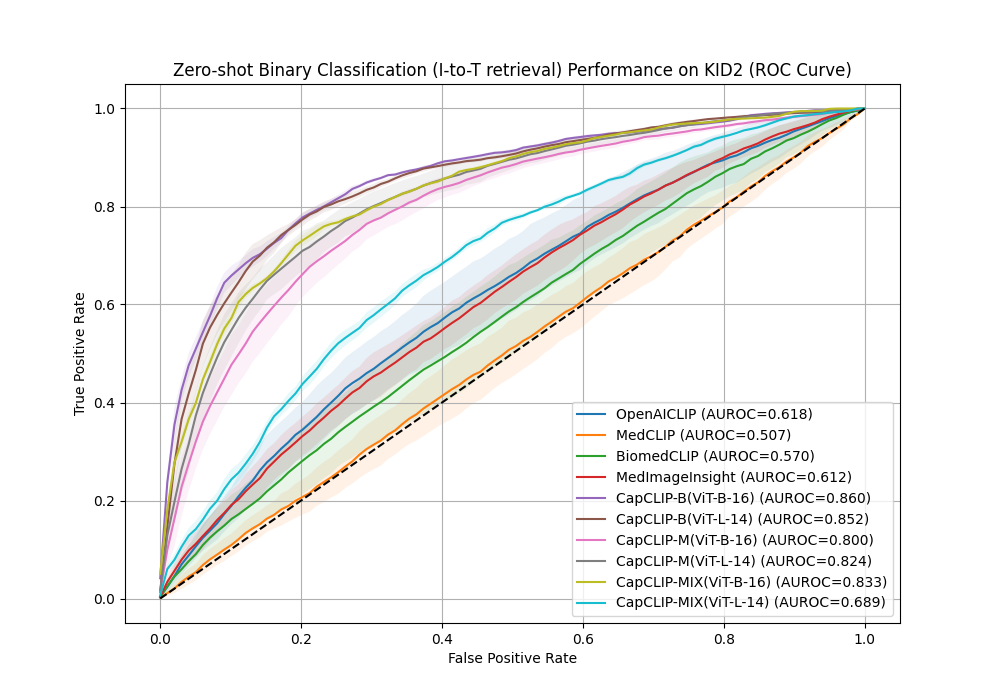}
        \caption{Mean ROC curve across different prompts for zero-shot CLIP-style binary classification on KID2.}
        \label{fig:kid2_roc}
    \end{subfigure}
    \hfill
    \begin{subfigure}[t]{0.7\textwidth}
        \centering
        \includegraphics[width=\linewidth]{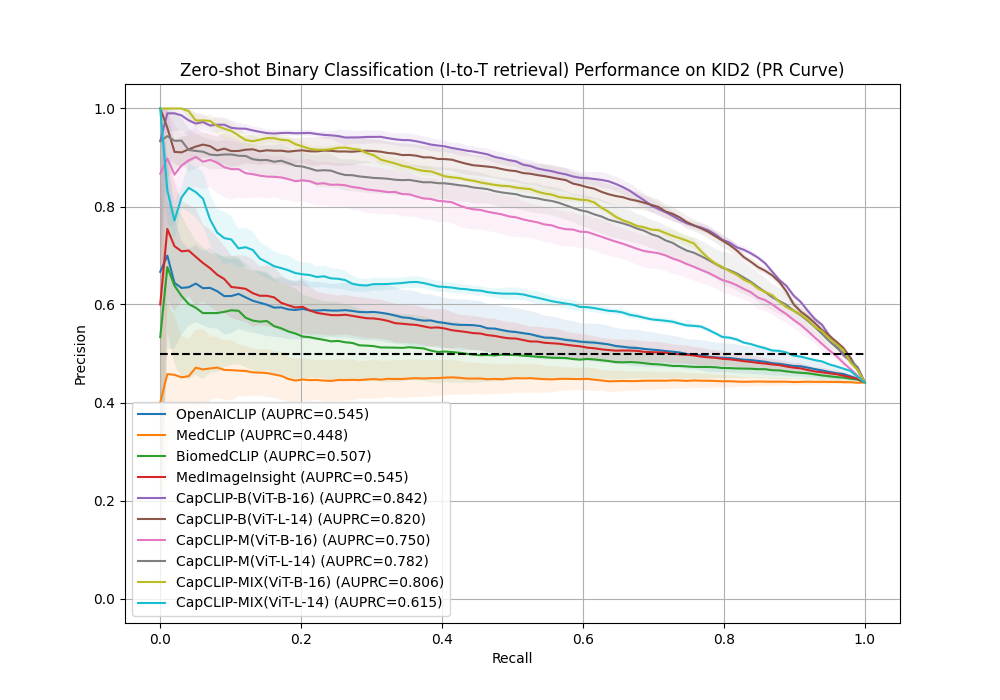}
        \caption{Mean precision-recall curve across different prompts for zero-shot CLIP-style binary classification on KID2.}
        \label{fig:kid2_pr}
    \end{subfigure}
    \caption{Zero-shot CLIP-style binary classification performance on the KID2 dataset. The shaded band indicates variation with respect to prompt sets, and the dotted horizontal line denotes the random baseline.}
    \label{fig:kid2_curves}
\end{figure*}

\begin{figure*}[t]
    \centering
    \begin{subfigure}[t]{0.7\textwidth}
        \centering
        \includegraphics[width=\linewidth]{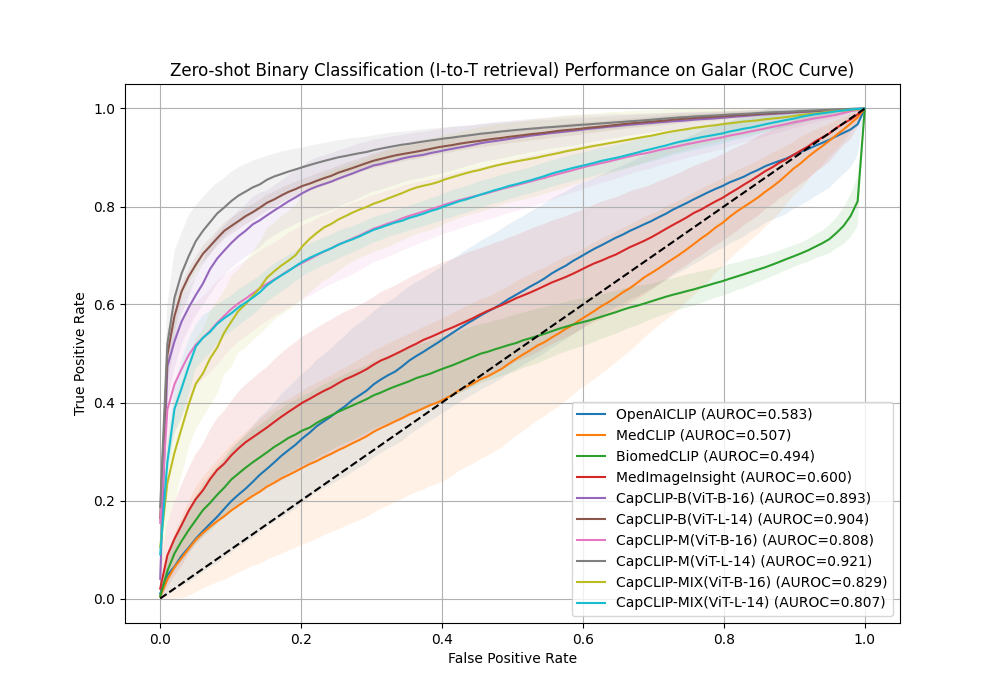}
        \caption{Mean ROC curve across different prompts for zero-shot CLIP-style binary classification on Galar.}
        \label{fig:galar_roc}
    \end{subfigure}
    \hfill
    \begin{subfigure}[t]{0.7\textwidth}
        \centering
        \includegraphics[width=\linewidth]{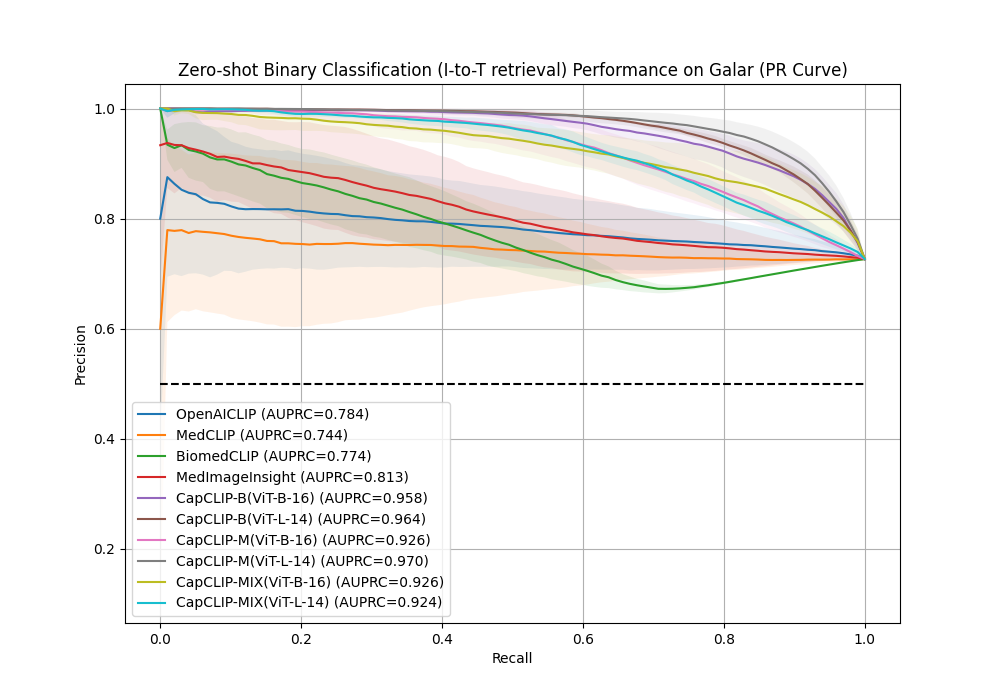}
        \caption{Mean precision-recall curve across different prompts for zero-shot CLIP-style binary classification on Galar.}
        \label{fig:galar_pr}
    \end{subfigure}
    \caption{Zero-shot CLIP-style binary classification performance on the Galar dataset. The shaded band indicates variation with respect to prompt sets, and the dotted horizontal line denotes the random baseline.}
    \label{fig:galar_curves}
\end{figure*}

This appendix presents supplementary results for zero-shot CLIP-style image-text classification. In particular, ROC and precision-recall curves for the datasets not shown in the main text are included here to complement the summary results reported in Section~V. Together with the CrohnIPI curves shown in the main paper, these figures provide a more complete view of binary image-text classification behaviour across datasets and prompt sets. KID2 and Galar ROC and precision-recall curves are provided in Figs.~\ref{fig:kid2_curves} and \ref{fig:galar_curves}, respectively.

\section{Detailed Text-to-Image Retrieval Results}
\label{sec:appendix_retrieval}

This appendix provides the detailed zero-shot text-to-image retrieval results corresponding to the retrieval summaries reported in the main text. Abnormality-level coarse retrieval results for KID2, CrohnIPI, and Galar are reported in Tables~\ref{tab:appendix_retrieval_coarse_kid2}--\ref{tab:appendix_retrieval_coarse_galar}. Pathology-level fine-grained retrieval results for the same datasets are reported in Tables~\ref{tab:appendix_retrieval_fine_kid2}--\ref{tab:appendix_retrieval_fine_galar}. These tables include the full mAP, Precision@\(K\), and Recall@\(K\) metrics used to support the retrieval analysis in Section~V.

\begin{table*}[tbp]
\centering
\caption{Zero-shot abnormality-level coarse text-to-image retrieval results for the KID2 dataset. The best baseline in each column is marked with an asterisk (*). Percentage improvement over the best baseline is shown in parentheses for the best CapCLIP model.}
\label{tab:appendix_retrieval_coarse_kid2}
\renewcommand{\arraystretch}{1.1}
\scriptsize
\resizebox{\textwidth}{!}{%
\begin{tabular}{lccccccc}
\hline
\textbf{Model} & \textbf{mAP} &
\multicolumn{3}{c}{\textbf{Precision@\(K\)}} &
\multicolumn{3}{c}{\textbf{Recall@\(K\) (support = 574)}} \\
\cline{3-5} \cline{6-8}
& & \textbf{\(K=100\)} & \textbf{\(K=200\)} & \textbf{\(K=500\)}
& \textbf{\(K=100\)} & \textbf{\(K=200\)} & \textbf{\(K=500\)} \\
\hline
CLIP-OpenAI         & 0.438  & 0.405  & 0.440  & 0.434  & 0.071  & 0.153  & 0.378  \\
MedCLIP             & 0.428  & 0.434  & 0.449* & 0.403  & 0.076  & 0.156* & 0.351  \\
BioMedCLIP          & 0.437  & 0.448  & 0.435  & 0.433  & 0.078  & 0.152  & 0.377  \\
MedImageInsight     & 0.461* & 0.478* & 0.448  & 0.451* & 0.083* & 0.156* & 0.393* \\
CapCLIP-B (ViT-B/16)   & 0.537 & 0.492 & 0.502 & 0.494 & 0.086 & 0.175 & 0.430 \\
CapCLIP-B (ViT-L/14)   & 0.636 (+38.0\%) & 0.666 (+39.3\%) & 0.662 (+47.4\%) & 0.631 (+39.9\%) & 0.116 (+39.8\%) & 0.231 (+48.1\%) & 0.550 (+39.9\%) \\
CapCLIP-M (ViT-B/16)   & 0.322 & 0.187 & 0.195 & 0.240 & 0.033 & 0.068 & 0.209 \\
CapCLIP-M (ViT-L/14)   & 0.325 & 0.207 & 0.195 & 0.231 & 0.036 & 0.068 & 0.201 \\
CapCLIP-MIX (ViT-B/16) & 0.538 & 0.527 & 0.541 & 0.542 & 0.092 & 0.188 & 0.472 \\
CapCLIP-MIX (ViT-L/14) & 0.540 & 0.597 & 0.566 & 0.521 & 0.104 & 0.197 & 0.454 \\
\hline
\end{tabular}%
}
\end{table*}

\begin{table*}[tbp]
\centering
\caption{Zero-shot abnormality-level coarse text-to-image retrieval results for the CrohnIPI dataset. The best baseline in each column is marked with an asterisk (*). Percentage improvement over the best baseline is shown in parentheses for the best CapCLIP model.}
\label{tab:appendix_retrieval_coarse_crohnipi}
\renewcommand{\arraystretch}{1.1}
\scriptsize
\resizebox{\textwidth}{!}{%
\begin{tabular}{lccccccc}
\hline
\textbf{Model} & \textbf{mAP} &
\multicolumn{3}{c}{\textbf{Precision@\(K\)}} &
\multicolumn{3}{c}{\textbf{Recall@\(K\) (support = 1360)}} \\
\cline{3-5} \cline{6-8}
& & \textbf{\(K=100\)} & \textbf{\(K=200\)} & \textbf{\(K=500\)}
& \textbf{\(K=100\)} & \textbf{\(K=200\)} & \textbf{\(K=500\)} \\
\hline
CLIP-OpenAI         & 0.448* & 0.511* & 0.493* & 0.482* & 0.038* & 0.073* & 0.177* \\
MedCLIP             & 0.356  & 0.274  & 0.324  & 0.333  & 0.020  & 0.048  & 0.122  \\
BioMedCLIP          & 0.421  & 0.320  & 0.375  & 0.431  & 0.024  & 0.055  & 0.159  \\
MedImageInsight     & 0.394  & 0.292  & 0.321  & 0.365  & 0.021  & 0.047  & 0.134  \\
CapCLIP-B (ViT-B/16)   & 0.774 & 0.779 & 0.786 & 0.787 & 0.057 & 0.116 & 0.290 \\
CapCLIP-B (ViT-L/14)   & 0.870 (+94.2\%) & 0.921 (+80.2\%) & 0.914 (+85.4\%) & 0.897 (+86.1\%) & 0.068 (+78.9\%) & 0.134 (+83.6\%) & 0.330 (+86.4\%) \\
CapCLIP-M (ViT-B/16)   & 0.304 & 0.161 & 0.161 & 0.174 & 0.012 & 0.024 & 0.064 \\
CapCLIP-M (ViT-L/14)   & 0.247 & 0.052 & 0.058 & 0.064 & 0.004 & 0.009 & 0.024 \\
CapCLIP-MIX (ViT-B/16) & 0.717 & 0.749 & 0.749 & 0.738 & 0.055 & 0.110 & 0.271 \\
CapCLIP-MIX (ViT-L/14) & 0.786 & 0.722 & 0.781 & 0.820 & 0.053 & 0.115 & 0.302 \\
\hline
\end{tabular}%
}
\end{table*}

\begin{table*}[tbp]
\centering
\caption{Zero-shot abnormality-level coarse text-to-image retrieval results for the Galar dataset. The best baseline in each column is marked with an asterisk (*). Percentage improvement over the best baseline is shown in parentheses for the best CapCLIP model where provided in the source table.}
\label{tab:appendix_retrieval_coarse_galar}
\renewcommand{\arraystretch}{1.1}
\scriptsize
\resizebox{\textwidth}{!}{%
\begin{tabular}{lccccccc}
\hline
\textbf{Model} & \textbf{mAP} &
\multicolumn{3}{c}{\textbf{Precision@\(K\)}} &
\multicolumn{3}{c}{\textbf{Recall@\(K\) (support = 2683)}} \\
\cline{3-5} \cline{6-8}
& & \textbf{\(K=100\)} & \textbf{\(K=200\)} & \textbf{\(K=500\)}
& \textbf{\(K=100\)} & \textbf{\(K=200\)} & \textbf{\(K=500\)} \\
\hline
CLIP-OpenAI         & 0.738  & 0.718  & 0.715  & 0.748  & 0.027  & 0.053  & 0.139  \\
MedCLIP             & 0.638  & 0.550  & 0.570  & 0.520  & 0.021  & 0.043  & 0.097  \\
BioMedCLIP          & 0.779* & 0.945* & 0.930* & 0.888* & 0.035* & 0.069* & 0.166* \\
MedImageInsight     & 0.697  & 0.629  & 0.621  & 0.650  & 0.024  & 0.046  & 0.121  \\
CapCLIP-B (ViT-B/16)   & 0.951 & 0.995 & 0.997 & 0.996 & 0.037 (+5.7\%) & 0.074 & 0.186 (+12.0\%) \\
CapCLIP-B (ViT-L/14)   & 0.966 (+24.0\%) & 0.991 & 0.995 & 0.998 (+12.4\%) & 0.037 (+5.7\%) & 0.074 & 0.186 (+12.0\%) \\
CapCLIP-M (ViT-B/16)   & 0.774 & 0.867 & 0.853 & 0.830 & 0.032 & 0.064 & 0.155 \\
CapCLIP-M (ViT-L/14)   & 0.665 & 0.603 & 0.591 & 0.604 & 0.023 & 0.044 & 0.113 \\
CapCLIP-MIX (ViT-B/16) & 0.924 & 1.000 (+5.8\%) & 0.999 (+7.4\%) & 0.997 & 0.037 (+5.7\%) & 0.075 (+8.7\%) & 0.186 (+12.0\%) \\
CapCLIP-MIX (ViT-L/14) & 0.930 & 0.994 & 0.995 & 0.990 & 0.037 (+5.7\%) & 0.074 & 0.185 \\
\hline
\end{tabular}%
}
\end{table*}
\begin{table*}[tbp]
\centering
\caption{Zero-shot pathology-level fine-grained text-to-image retrieval results for the KID2 dataset. Macro and weighted average mAP and Precision@\(K\) are reported. The best baseline in each column is marked with an asterisk (*). Percentage improvement over the best baseline is shown in parentheses for the best CapCLIP model.}
\label{tab:appendix_retrieval_fine_kid2}
\renewcommand{\arraystretch}{1.1}
\scriptsize
\resizebox{\textwidth}{!}{%
\begin{tabular}{lcccccccc}
\hline
\textbf{Model} &
\textbf{mAP} &
\textbf{mAP} &
\multicolumn{3}{c}{\textbf{Macro-average Precision@\(K\)}} &
\multicolumn{3}{c}{\textbf{Weighted-average Precision@\(K\)}} \\
&
\textbf{Macro} &
\textbf{Weighted} &
\textbf{\(K=100\)} & \textbf{\(K=200\)} & \textbf{\(K=500\)} &
\textbf{\(K=100\)} & \textbf{\(K=200\)} & \textbf{\(K=500\)} \\
\hline
CLIP-OpenAI         & 0.177 & 0.221 & 0.181  & 0.175  & 0.166  & 0.226  & 0.225* & 0.214 \\
MedCLIP             & 0.148 & 0.189 & 0.143  & 0.151  & 0.134  & 0.175  & 0.189  & 0.172 \\
BioMedCLIP          & 0.184 & 0.222 & 0.199* & 0.184* & 0.172* & 0.231  & 0.223  & 0.219 \\
MedImageInsight     & 0.194* & 0.233* & 0.191 & 0.180 & 0.171 & 0.234* & 0.223 & 0.222* \\
CapCLIP-B (ViT-B/16)   & 0.191 & 0.255 & 0.195 & 0.176 & 0.166 & 0.256 & 0.239 & 0.227 \\
CapCLIP-B (ViT-L/14)   & 0.232 & 0.302 & 0.251 & 0.240 & 0.226 & 0.329 & 0.313 & 0.295 \\
CapCLIP-M (ViT-B/16)   & 0.219 & 0.289 & 0.227 & 0.225 & 0.218 & 0.304 & 0.303 & 0.289 \\
CapCLIP-M (ViT-L/14)   & 0.244 & 0.310 & 0.293 & 0.266 & 0.220 & 0.379 & 0.340 & 0.280 \\
CapCLIP-MIX (ViT-B/16) & 0.316 & 0.408 & 0.369 & 0.335 & 0.253 & 0.481 & 0.438 & 0.332 \\
CapCLIP-MIX (ViT-L/14) & 0.385 (+98.5\%) & 0.467 (+100.4\%) & 0.468 (+135.2\%) & 0.376 (+104.3\%) & 0.254 (+47.7\%) & 0.583 (+149.1\%) & 0.479 (+112.9\%) & 0.332 (+49.5\%) \\
\hline
\end{tabular}%
}
\end{table*}

\begin{table*}[tbp]
\centering
\caption{Zero-shot pathology-level fine-grained text-to-image retrieval results for the CrohnIPI dataset. Macro and weighted average mAP and Precision@\(K\) are reported. The best baseline in each column is marked with an asterisk (*). Percentage improvement over the best baseline is shown in parentheses for the best CapCLIP model.}
\label{tab:appendix_retrieval_fine_crohnipi}
\renewcommand{\arraystretch}{1.1}
\scriptsize
\resizebox{\textwidth}{!}{%
\begin{tabular}{lcccccccc}
\hline
\textbf{Model} &
\textbf{mAP} &
\textbf{mAP} &
\multicolumn{3}{c}{\textbf{Macro-average Precision@\(K\)}} &
\multicolumn{3}{c}{\textbf{Weighted-average Precision@\(K\)}} \\
&
\textbf{Macro} &
\textbf{Weighted} &
\textbf{\(K=100\)} & \textbf{\(K=200\)} & \textbf{\(K=500\)} &
\textbf{\(K=100\)} & \textbf{\(K=200\)} & \textbf{\(K=500\)} \\
\hline
CLIP-OpenAI         & 0.080* & 0.093* & 0.080* & 0.083* & 0.079  & 0.097* & 0.101* & 0.095  \\
MedCLIP             & 0.062  & 0.079  & 0.045  & 0.053  & 0.055  & 0.062  & 0.071  & 0.075  \\
BioMedCLIP          & 0.078  & 0.093* & 0.073  & 0.082  & 0.085* & 0.085  & 0.093  & 0.100* \\
MedImageInsight     & 0.066  & 0.079  & 0.035  & 0.038  & 0.050  & 0.040  & 0.045  & 0.060  \\
CapCLIP-B (ViT-B/16)   & 0.107 & 0.119 & 0.090 & 0.097 & 0.092 & 0.102 & 0.107 & 0.102 \\
CapCLIP-B (ViT-L/14)   & 0.158 & 0.182 & 0.150 & 0.145 & 0.145 & 0.177 & 0.173 & 0.170 \\
CapCLIP-M (ViT-B/16)   & 0.098 & 0.102 & 0.085 & 0.088 & 0.079 & 0.085 & 0.088 & 0.081 \\
CapCLIP-M (ViT-L/14)   & 0.071 & 0.071 & 0.051 & 0.048 & 0.045 & 0.049 & 0.045 & 0.043 \\
CapCLIP-MIX (ViT-B/16) & 0.171 & 0.195 & 0.155 & 0.147 & 0.155 & 0.174 & 0.168 & 0.181 \\
CapCLIP-MIX (ViT-L/14) & 0.225 (+181.3\%) & 0.230 (+147.3\%) & 0.219 (+173.8\%) & 0.215 (+159.0\%) & 0.184 (+116.5\%) & 0.214 (+120.6\%) & 0.216 (+113.9\%) & 0.209 (+109.0\%) \\
\hline
\end{tabular}%
}
\end{table*}

\begin{table*}[tbp]
\centering
\caption{Zero-shot pathology-level fine-grained text-to-image retrieval results for the Galar dataset. Macro and weighted average mAP and Precision@\(K\) are reported. The best baseline in each column is marked with an asterisk (*). Percentage improvement over the best baseline is shown in parentheses for the best CapCLIP model.}
\label{tab:appendix_retrieval_fine_galar}
\renewcommand{\arraystretch}{1.1}
\scriptsize
\resizebox{\textwidth}{!}{%
\begin{tabular}{lcccccccc}
\hline
\textbf{Model} &
\textbf{mAP} &
\textbf{mAP} &
\multicolumn{3}{c}{\textbf{Macro-average Precision@\(K\)}} &
\multicolumn{3}{c}{\textbf{Weighted-average Precision@\(K\)}} \\
&
\textbf{Macro} &
\textbf{Weighted} &
\textbf{\(K=100\)} & \textbf{\(K=200\)} & \textbf{\(K=500\)} &
\textbf{\(K=100\)} & \textbf{\(K=200\)} & \textbf{\(K=500\)} \\
\hline
CLIP-OpenAI         & 0.087  & 0.104  & 0.083  & 0.077  & 0.073  & 0.073  & 0.073  & 0.073 \\
MedCLIP             & 0.065  & 0.082  & 0.050  & 0.052  & 0.045  & 0.048  & 0.049  & 0.044 \\
BioMedCLIP          & 0.102* & 0.118* & 0.102* & 0.099* & 0.092* & 0.099* & 0.097* & 0.089* \\
MedImageInsight     & 0.085  & 0.101  & 0.079  & 0.078  & 0.073  & 0.071  & 0.072  & 0.070 \\
CapCLIP-B (ViT-B/16)   & 0.118 & 0.254 & 0.090 & 0.089 & 0.080 & 0.264 & 0.262 & 0.226 \\
CapCLIP-B (ViT-L/14)   & 0.113 & 0.229 & 0.076 & 0.081 & 0.087 & 0.187 & 0.199 & 0.212 \\
CapCLIP-M (ViT-B/16)   & 0.128 & 0.149 & 0.119 & 0.122 & 0.112 & 0.118 & 0.126 & 0.118 \\
CapCLIP-M (ViT-L/14)   & 0.176 (+72.5\%) & 0.189 & 0.190 (+86.3\%) & 0.166 (+67.7\%) & 0.137 (+48.9\%) & 0.211 & 0.192 & 0.166 \\
CapCLIP-MIX (ViT-B/16) & 0.141 & 0.260 & 0.156 & 0.149 & 0.119 & 0.305 & 0.304 & 0.251 (+182.0\%) \\
CapCLIP-MIX (ViT-L/14) & 0.155 & 0.280 (+137.3\%) & 0.176 & 0.166 (+67.7\%) & 0.118 & 0.361 (+264.6\%) & 0.350 (+260.8\%) & 0.246 \\
\hline
\end{tabular}%
}
\end{table*}

\FloatBarrier

\clearpage

\bibliographystyle{IEEEtran}
\bibliography{references}

\end{document}